%% file: main.tex
\documentclass[10pt,twocolumn,letterpaper]{article}

\makeatletter
\@namedef{ver@everyshi.sty}{}
\makeatother

\usepackage[pagenumbers]{cvpr} %

\usepackage{graphicx}
\usepackage{amsmath}
\usepackage{amssymb}
\usepackage{booktabs}

\usepackage{comment}
\usepackage{color}
\usepackage{microtype}
\usepackage[dvipsnames,table]{xcolor}
\usepackage{tikz}
\usepackage[framemethod=tikz]{mdframed}
\usepackage{tabularx}
\usepackage{makecell}
\usepackage{wrapfig}
\usepackage{multirow}
\usepackage{ctable}
\usepackage{pifont}
\usepackage[multiple]{footmisc}
\usepackage[shortlabels]{enumitem}
\usepackage[T1]{fontenc}
\usepackage[utf8]{inputenc}
\usepackage{pmboxdraw}

\usepackage{listings}
\usepackage[numbers,sort&compress]{natbib}

\newtheorem{lemma}{Lemma}

\usepackage{algorithm}
\usepackage[noend]{algpseudocode}

\input{math_commands}

\newcommand*{\ShowNotes}{}
\input{macros.tex}

\usepackage{symbols}
\usepackage[makeroom]{cancel}

\usepackage{physics}
\newcommand\scalemath[2]{\scalebox{#1}{\mbox{\ensuremath{\displaystyle #2}}}}

\newcommand{\gauss}[2]{\mathcal{N}(#1,\, #2)}
\newcommand{\score}[2]{\grad_{#1} \log p_{#2}(#1)}
\newenvironment{proof}{\vspace{-0.2cm}{\it Proof.} }{\hfill$\square$}

\newcommand{\vomega}{\bm{\omega}}

\definecolor{mybrown}{HTML}{c25b14}
\definecolor{Ddotcol}{RGB}{68, 114, 196}

\newcommand{\escore}{Perturb-and-Average Scoring\xspace}
\newcommand{\nescore}{\text{PAAS}\xspace}

\usepackage{tikz-cd}

\usepackage[pagebackref,breaklinks,colorlinks]{hyperref}

\usepackage[capitalize]{cleveref}
\crefname{section}{Sec.}{Secs.}
\Crefname{section}{Section}{Sections}
\Crefname{table}{Table}{Tables}
\crefname{table}{Tab.}{Tabs.}

\usepackage{thm-restate}
\mdfdefinestyle{MyFrame}{%
    linecolor=Black,
    outerlinewidth=0.1pt,
    roundcorner=2pt,
    innertopmargin=0.3\baselineskip,
    innerbottommargin=0.3\baselineskip,
    innermargin = 0.2cm,
  	outermargin = 0.cm,
    innerrightmargin=5pt,
    innerleftmargin=5pt,
    backgroundcolor=Gray!10!white}
    
\def\cvprPaperID{3570} %
\def\confName{CVPR}
\def\confYear{2023}

\begin{document}

\title{Score Jacobian Chaining: Lifting Pretrained 2D Diffusion Models\\ for 3D Generation} 

\author{
    Haochen Wang$^{*1}$ \quad Xiaodan Du$^{*1}$ \quad Jiahao Li$^{*1}$ \quad Raymond A. Yeh$^2$ \quad Greg Shakhnarovich$^1$ \\
    {\vspace{-0.5em}} \\
    $^1$TTI-Chicago $\quad \quad$ $^2$Purdue University
}

\twocolumn[{%
\renewcommand\twocolumn[1][]{#1}%
\maketitle
\begin{center}
    \vspace{-0.5cm}
    \centering
    \setlength{\tabcolsep}{3pt}
    \renewcommand{\arraystretch}{0.95}
    \begin{tabular}{cc}
        \multicolumn{1}{c}{\small A zoomed out high-quality photo of Temple of Heaven} & \multicolumn{1}{c}{\small A high quality photo of a delicious burger}\\
    \includegraphics[width=0.5\linewidth]{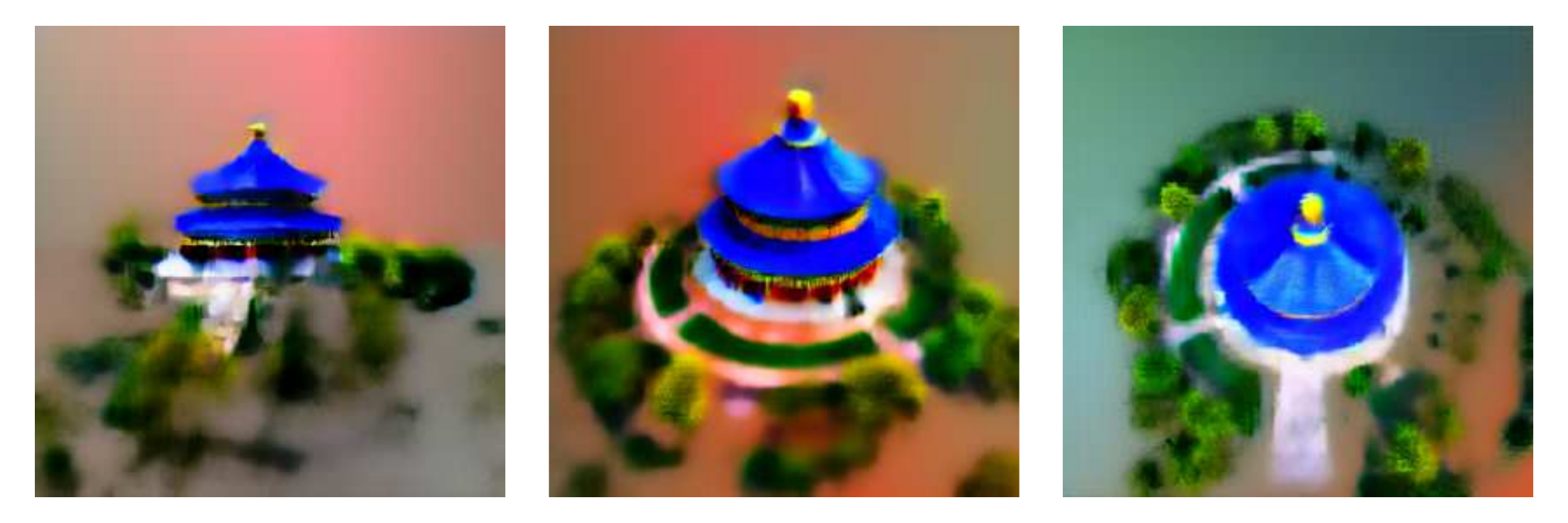} & 
    \includegraphics[width=0.5\linewidth]{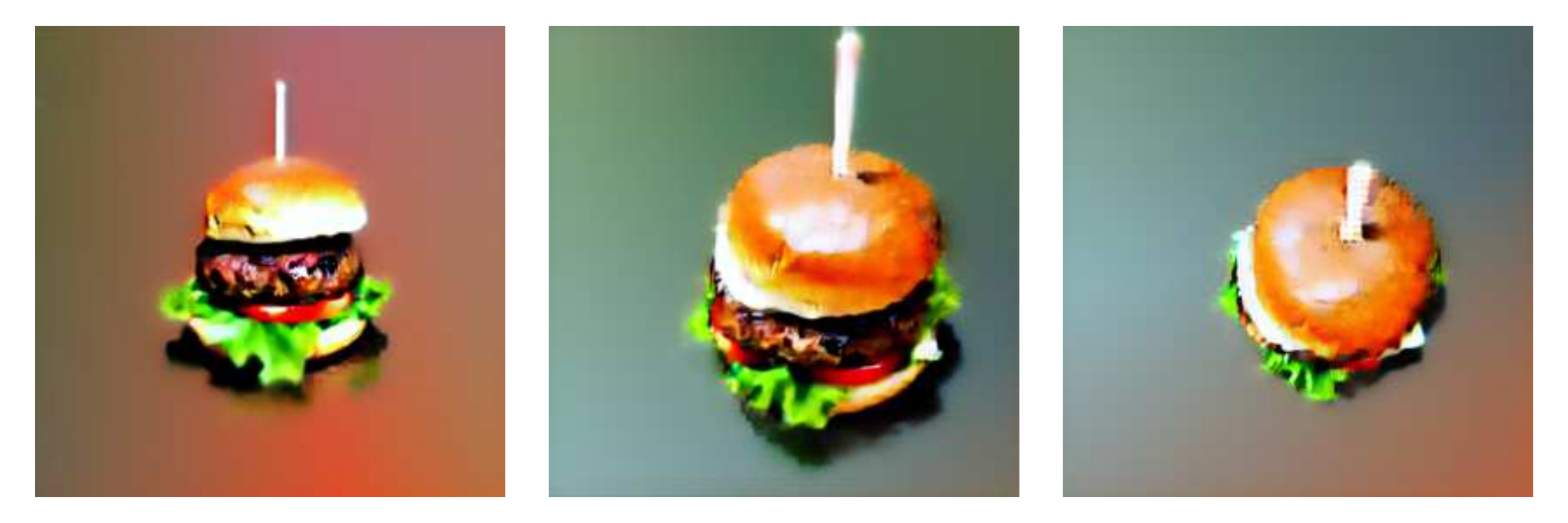} \\
    \multicolumn{1}{c}{\small 
    \begin{tabular}{l}
    A high quality photo of a Victorian style wooden chair\\ with velvet upholstery
    \end{tabular}} 
    & \multicolumn{1}{c}{\small A high quality photo of a classic silver muscle car}\\
    \includegraphics[width=0.5\linewidth]{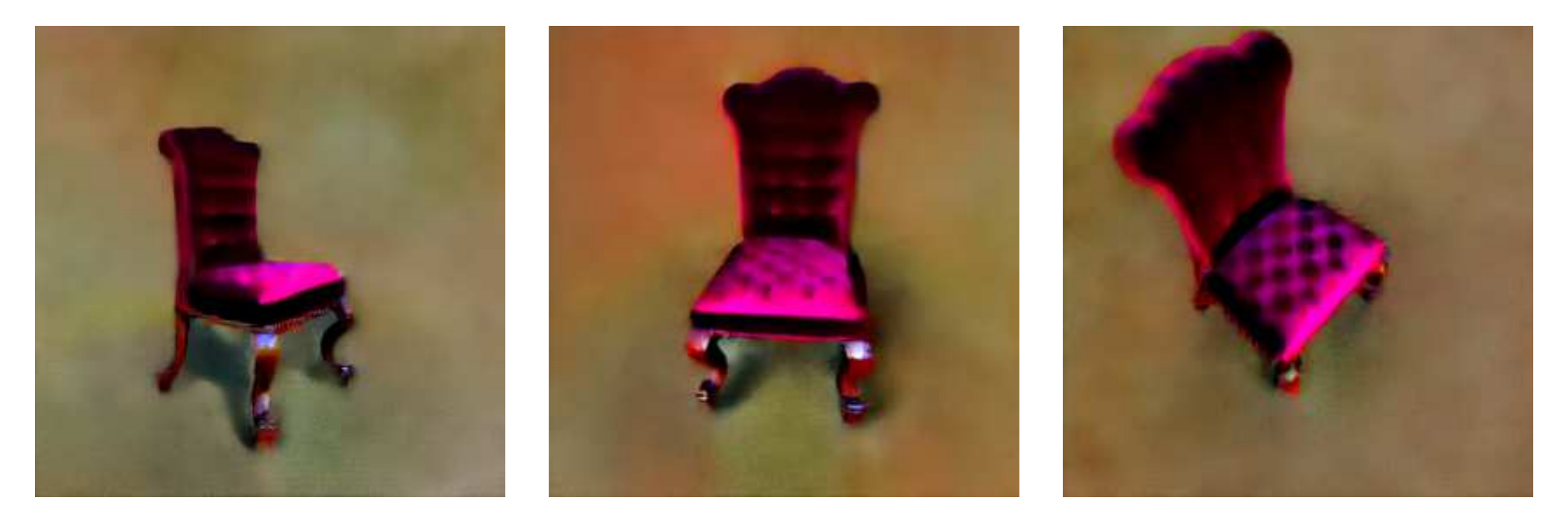} & 
    \includegraphics[width=0.5\linewidth]{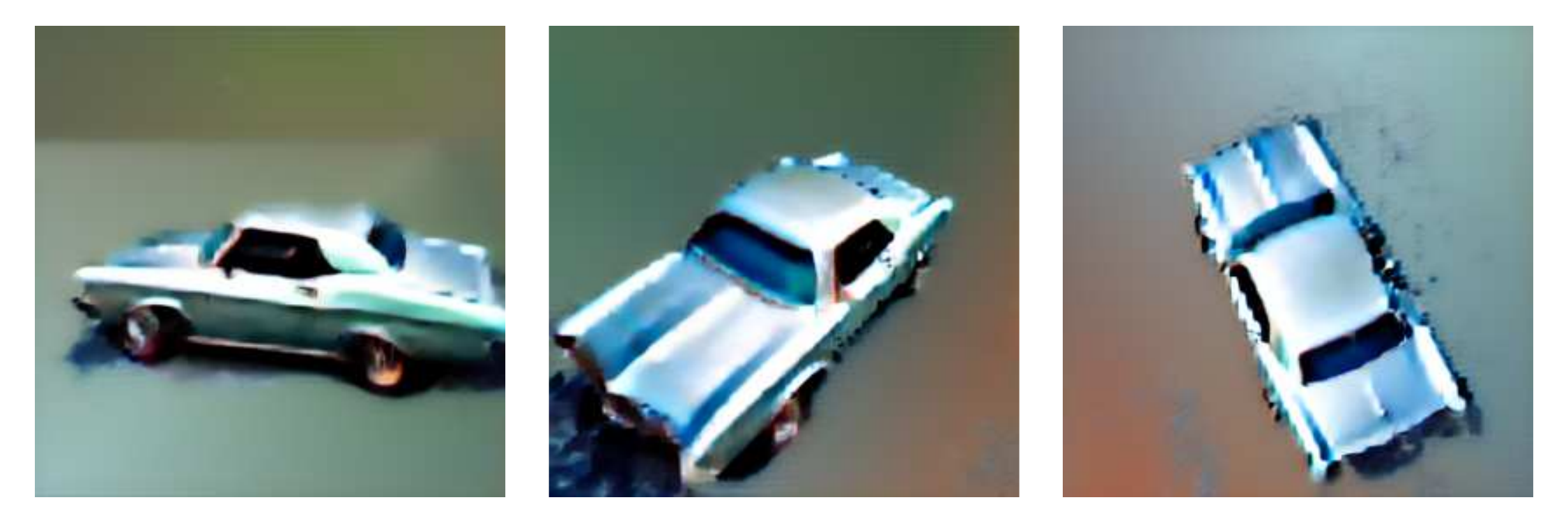} \\
    \end{tabular}
\end{center}%
    \vspace{-0.8cm}
    
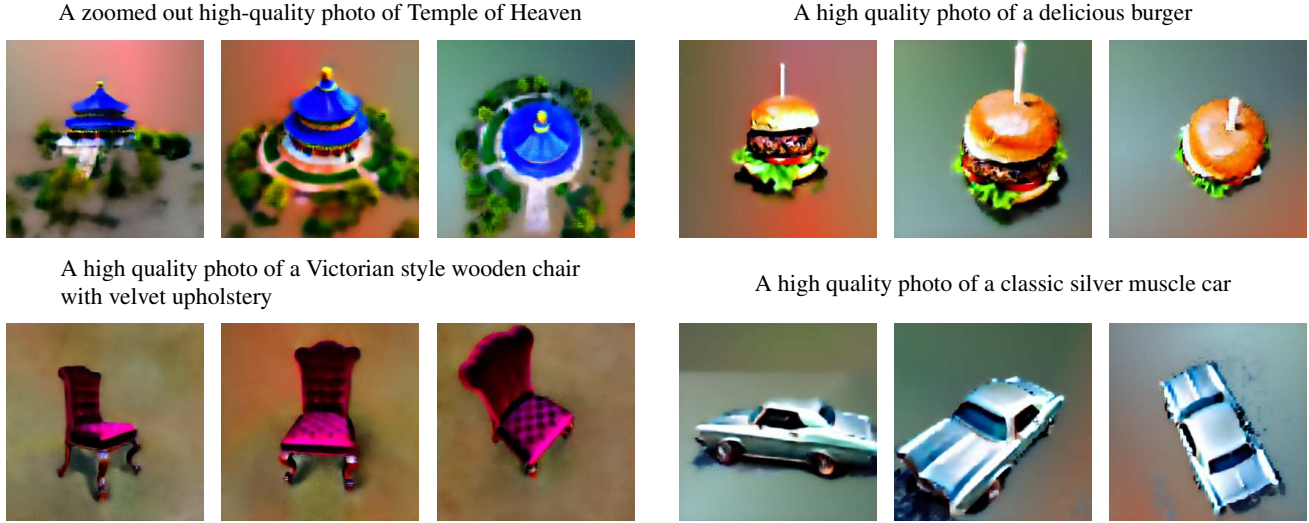
\captionof{figure}{Results for text-driven 3D generation using Score Jacobian Chaining with Stable Diffusion as the pretrained model.}
    \label{fig:teaser}
\vspace{0.3cm}
}]

\begin{abstract}
\input{00_abs}
\end{abstract}
\let\thefootnote\relax\footnote{* Equal contribution.}

\input{01_intro}

\input{02_rel}

\input{03_prelim}

\input{04_app}

\input{05_exp}

\input{06_conc}

{
\small
\bibliographystyle{ieee_fullname}
\bibliography{ref}
}

\clearpage
\input{07_appendix}

\end{document}

%% file: math_commands.tex
\usepackage{amsmath,amsfonts,bm}

\def\1{\bm{1}}

\def\sp{space}

\def\rmI{{\mathbf{I}}}

\def\vmu{{\bm{\mu}}}
\def\vtheta{{\bm{\theta}}}

\def\vn{{\bm{n}}}

\def\vx{{\bm{x}}}
\def\vy{{\bm{y}}}
\def\vz{{\bm{z}}}

\def\mD{{\bm{D}}}

\def\mI{{\bm{I}}}
\def\mJ{{\bm{J}}}

\DeclareMathAlphabet{\mathsfit}{\encodingdefault}{\sfdefault}{m}{sl}
\SetMathAlphabet{\mathsfit}{bold}{\encodingdefault}{\sfdefault}{bx}{n}

\def\gL{{\mathcal{L}}}

\def\gN{{\mathcal{N}}}

\def\gY{{\mathcal{Y}}}

\newcommand{\E}{\mathbb{E}}

\newcommand{\R}{\mathbb{R}}

\newcommand{\Var}{\mathrm{Var}}

%% file: macros.tex
\definecolor{darkred}{rgb}{0.7,0.1,0.1}
\definecolor{darkgreen}{rgb}{0.1,0.7,0.1}
\definecolor{cyan}{rgb}{0.7,0.0,0.7}
\definecolor{dblue}{rgb}{0.2,0.2,0.8}
\definecolor{maroon}{rgb}{0.76,.13,.28}
\definecolor{burntorange}{rgb}{0.81,.33,0}
\definecolor{tealblue}{rgb}{0.212,0.459, 0.533}

\definecolor{mypink}{rgb}{0.93359375, 0.62109375, 0.83984375}

\definecolor{pp}{rgb}{0.43921569, 0.18823529, 0.62745098}
\definecolor{rr}{rgb}{0.5254902 , 0.00784314, 0.12941176}
\definecolor{bb}{rgb}{0.09019608, 0.23529412, 0.37647059}
\definecolor{yy}{rgb}{0.49803922, 0.3372549 , 0.0}
\definecolor{gg}{rgb}{0.02352941, 0.3372549 , 0.17647059}

\ifdefined\ShowNotes
  \newcommand{\colornote}[3]{{\color{#1}\bf{#2: #3}\normalfont}}
\else
  \newcommand{\colornote}[3]{}
\fi

\newcommand{\eat}[1]{} %

%% file: 00_abs.tex
A diffusion model learns to predict a vector field of gradients. We propose to apply chain rule on the learned gradients, and back-propagate the score of a diffusion model through the Jacobian of a differentiable renderer, which we instantiate to be a voxel radiance field. This setup aggregates 2D scores at multiple camera viewpoints into a 3D score,
and repurposes a pretrained 2D model for 3D data generation. We identify a technical challenge of distribution mismatch that arises in this application, and propose a novel estimation mechanism to resolve it.
We run our algorithm on several off-the-shelf diffusion image generative models, including the recently released Stable Diffusion trained on the large-scale LAION 5B dataset. 

%% file: 01_intro.tex
\vspace{-1em}
\section{Introduction}

We introduce a method that converts a pretrained 2D diffusion generative model on images into a 3D generative model of radiance fields, without requiring access to any 3D data. The key insight is to interpret diffusion models as learned predictors of a gradient field, often referred to as the \emph{score function} of the data log-likelihood. We apply the chain rule on the estimated score, hence the name Score Jacobian Chaining (SJC). 

Following~\citet{hyvarinen2005estimation}, the score is defined as the gradient of the log-density function with respect to the data. Diffusion models of various families~\cite{song2019generative, sohl2015deep, ho2020denoising, song2021score} can all be interpreted~\cite{song2021score, kingma2021variational, karras2022elucidating} as modeling $\score{\vx}{\sigma}$ \ie the denoising score
at noise level $\sigma$. For readability, we refer to the denoising score as the score.
Generating a sample from a diffusion model involves repeated evaluations of the score function 
from large to small $\sigma$ level, so that a sample $\vx$ 
gradually moves closer to the data manifold. 
It can be loosely interpreted as gradient descent, with 
precise control on the step sizes so that data distribution evolves to match the annealed $\sigma$ level
(ancestral sampler~\cite{ho2020denoising}, SDE and probability-flow ODE~\cite{song2021score}, etc.). 
While there are other perspectives to a diffusion model~\cite{sohl2015deep,ho2020denoising}, here we are primarily motivated from the viewpoint that diffusion models produce a gradient field.

A natural question to ask is whether the chain rule can be applied to the learned gradients. Consider a diffusion model on images. An image $\vx$ may be parameterized by some function $f$ with parameters $\vtheta$, \ie, $\vx = f(\vtheta)$. Applying the chain rule through the Jacobian $\pdv{\vx}{\vtheta}$ converts a gradient on image $\vx$ into a gradient on the parameter $\vtheta$. There are many potential use cases for pairing a pretrained diffusion model with different choices of $f$. In this work we are interested in exploring the connection between 3D and multiview 2D by choosing $f$ to be a differentiable renderer, thus creating a 3D generative model using only pretrained 2D resources. 

Many prior works~\cite{pointflow, shapegf, wu3dgan} perform 3D generative modeling by training on 3D datasets~\cite{chang2015shapenet, sun2018pix3d, modelnet40, koch2019abc}. This approach is often as challenging as it is format-ambiguous. 
In addition to the high data acquisition cost of 3D assets~\cite{deltombe_2022}, there is no universal data format: point clouds, meshes, volumetric radiance field, etc, 
all have computational trade-offs. What is common to these 3D assets is that they can be rendered into 2D images. 
An inverse rendering system, or a differentiable renderer~\cite{opendr, pulsar, mitsuba2, li2018differentiable, nerf}, provides access to the Jacobian $\mJ_\pi \triangleq \pdv{\vx_{\pi}}{\vtheta}$ of a rendered image $\vx_{\pi}$ at camera viewpoint $\pi$ with respect to the underlying 3D parameterization $\vtheta$. Our method uses differentiable rendering to aggregate 2D image gradients over multiple viewpoints into a 3D asset gradient, and lifts a generative model from 2D to 3D. We parameterize a 3D asset $\vtheta$ as a radiance field stored on voxels and choose $f$ to be the volume rendering function. 

A key technical challenge is that computing the 2D score by directly evaluating a diffusion model on a rendered image $\vx_\pi$ leads to an out-of-distribution (OOD) problem. 
Generally, diffusion models are trained as denoisers and have only seen noisy inputs during training.
On the other hand, our method requires evaluating the denoiser on non-noisy rendered images from a 3D asset during optimization, and it leads to the OOD problem.
To address the issue, we propose \emph{\escore{}}, an approach to estimate the score for non-noisy images. 

Empirically, we first validate the effectiveness of \emph{\escore{}} at solving 
the OOD problem and explore the hyperparameter choices on a simple 2D image canvas. 
Here we identify open problems on using unconditioned diffusion models 
trained on FFHQ and LSUN Bedroom. 
Next, we use Stable Diffusion, a model pretrained on the web-scale LAION dataset to perform SJC for 3D generation, as shown in~\figref{fig:teaser}.
{\bf \noindent Our contributions are as follows:}
\begin{itemize}[topsep=3pt]%
    \setlength\itemsep{1pt}
    \item We propose a method for lifting a 2D diffusion model to 3D via an application of the chain rule.
    \item We illustrate the challenge of OOD when using a pretrained denoiser and propose \emph{\escore{}} to resolve it.
    \item We point out the subtleties and open problems on applying \emph{\escore{}} as gradient for optimization. 
    \item We demonstrate the effectiveness of SJC for the task of 3D text-driven generation.
\end{itemize}

%% file: 02_rel.tex
\section{Related Works}
{\noindent\bf Diffusion models} have
recently advanced to image generation on Internet-scale datasets~\cite{glide, dalle2, imagen, rombach2022high, laion, clip, dbeatgan}. A diffusion model 
could be interpreted as either a VAE~\cite{ho2020denoising, sohl2015deep} or a denoising score-matcher~\cite{vincent2011connection, song2019generative, song2021score}. Notably, models trained under one regime can be directly used
for inference and sampling by the other~\cite{song2021score, karras2022elucidating}; they are in practice largely equivalent. 

{\noindent\bf Neural radiance fields (NeRF)} is a family of inverse rendering algorithms that
have excelled at multiview 3D reconstruction tasks including view synthesis and surface geometry estimation~\cite{nerf, nerfw, unisurf, volsdf, neus}. Conceptually, a 3D asset is represented as a dense grid of $\mathrm{RGB}$ colors and spatial density $\tau$, and rendered into images in a way analogous to alpha compositing~\cite{maxvol}. 
NeRF parameterizes the $(\mathrm{RGB},\tau)$ volume with a neural network, but querying 
the network densely in 3D incurs significant compute costs. 
Alternatively, Voxel NeRFs~\cite{nsvf, dvgo, plenoxels, tensorf} store the volume on voxels and observe no loss in end task performance~\cite{dvgo, plenoxels}. Querying voxels is a simple 
memory operation that is much faster than a feedforward pass of a neural network. Here we use a customized voxel radiance field with hyperparameters based on DvGO~\cite{dvgo} and TensoRF~\cite{tensorf}.

\smallskip
{\noindent\bf 2D-supervised 3D GANs} pioneered~\cite{nguyen2019hologan,pix2scene,zhang2020image} the approach of training 3D generative models using only unstructured 2D images, and promise greater scalability in terms of data. Rather than supervising directly on the 3D asset a model generates, these methods supervise the 2D renderings of the generated 3D asset, often using an adversarial loss~\cite{graf, pigan, giraffe, eg3d, zhao2022generative}. In other words, only images are needed as training data. However, training such a 3D generative model from scratch is still challenging~\cite{campari}. Recent empirical evaluation remains mostly on human and animal faces~\cite{eg3d}. Our method does the opposite: we take an image generative model that is \textit{already pretrained} on large amounts of 2D data and use it to guide the iterative optimization of a 3D asset. Optimization-based generation makes it much slower compared to 3D GANs, but it becomes possible to harness powerful off-the-shelf 2D generative models such as Stable Diffusion~\cite{rombach2022high} for greater content diversity. 

\smallskip
{\noindent\bf CLIP-guided, optimization-based 3D generative models} share a similar philosophy
of optimizing 3D assets by guiding on 2D renderings~\cite{jetchev2021clipmatrix, michel2022text2mesh,clipmesh,dreamfields,clipvrf,avatarclip}. Among them, DreamFields~\cite{dreamfields} and PureClipNeRF~\cite{clipvrf} also use NeRF as their differentiable renderers. 
In this case, the 2D guidance comes from CLIP~\cite{clip}, 
a pretrained image-text matching model. These works optimize the 3D assets so that the 
image renderings match a user-provided text prompt. Since CLIP is not a 2D generative
model per se, such a pipeline usually creates some abstract distilled content~\cite{fusedream} 
that looks very different from real images. 
In contrast, we use diffusion models, which are proper 2D generative models, to create realistic looking 3D content. 

\smallskip
{\noindent\bf DreamFusion}. The recently arXived work by~\citet{dreamfusion}, \textit{independent and concurrent to our work}, proposes an algorithm that is similar to our approach at the pseudo-code level. Differently, their procedure uses the mathematical setup by~\citet{ppprior} to
search for image parametrization that minimizes the training loss of a diffusion model. 
In contrast, our work is motivated by applying the chain rule to the 2D score. The key differences have been summarized in~\secref{sec:compare_dream_fusion}. In terms of implementation, we do not have access to the close-sourced Imagen~\cite{imagen} diffusion model. Instead, we use the pretrained Stable Diffusion model released by~\citet{rombach2022high}. 
For a comparison with DreamFusion, we use with a third-party implementation based on the same diffusion model, namely Stable-DreamFusion\footnote{\small\text{\url{github.com/ashawkey/stable-dreamfusion}}}.

%% file: 03_prelim.tex
\section{Preliminaries}
To establish a common notation, we briefly review the score-based perspective of diffusion models. For readers familiar with VAE literature on diffusion models, we provide a concise score-based formula card and more details in Appendix~\secref{sec:score_perspective} to connect these ideas. 

\smallskip
{\bf\noindent Denoising score matching.}
Given a dataset of samples $\gY=\{\vy_i\}$ drawn from $p_{\text{data}}$, 
a diffusion model revolves primarily around learning a denoiser $D$
by minimizing the difference between a noised sample $\vy+ \sigma\vn$ and $\vy$, 
\bea\label{eq:training}
\E_{\vy \sim p_{\text{data}}}\E_{\vn \sim \gN(0, \mI)} \|D(\vy+\sigma\vn; \sigma)-\vy\|_2^2,
\eea
\ie $D$ is denoising the input $\vy+\sigma\vn$, for a range of $\sigma$ values. For 2D images, $D$ is commonly chosen to be a ConvNet. Variants such as DDPM~\cite{ho2020denoising} parameterized the ConvNet to instead predict
a noise residual $\hat{\bm{\epsilon}}$, and these models can be converted back to the form of
a denoiser by~\cite{song2021score} 
\bea
D(\vx; \sigma) = \vx - \sigma \hat{\bm{\epsilon}}(\vx).
\eea
In this paper, we treat all pretrained diffusion models as denoisers, and perform the interface
conversion in our implementation when needed. 

\smallskip
{\bf\noindent Score from denoiser.}
Let $p_{\sigma}(\vx)$ denote the data distribution perturbed by Gaussian noise of standard deviation $\sigma$. It is shown in prior works~\cite{song2019generative, hyvarinen2005estimation} that the denoiser $D$ trained according to \equref{eq:training} provides a good approximation to the denoising score:
\bea \label{eq:score_approx}
\score{\vx}{\sigma} \approx \frac{D(\vx; \sigma) -\vx}{\sigma^2}.
\eea 
A denoising diffusion model estimates the \textit{score function} of the noised distribution $p_\sigma(\vx)$ at various $\sigma \in \{\sigma_i\}_{i=1}^T$. To perform sampling, the diffusion model gradually updates a sample through a sequence of noise levels of $\sigma_T > \dots > \sigma_0 = 0$. $\{\sigma_i\}$ are chosen empirically, with a typical range being $[0.01, 157]$~\cite{ho2020denoising} in the case of DDPM. 

\smallskip
{\bf\noindent Score as mean-shift.} A helpful intuition is that the score behaves like \emph{mean-shift}~\cite{meanshift,mean-shift}. If we simplify $p_{\text{data}}$ to be an empirical data distribution over the i.i.d. samples $\{\vy_i\}$, then at noise level $\sigma$, $p_{\sigma}(\vx)$ takes the form of a mixture of Gaussians~\cite{song2020improved}
\begin{align}\label{eq:gmm}
p_{\sigma}(\vx) = \E_{\vy \sim p_\text{data}} ~ \mathcal{N}(\vx;\, \vy, \, \sigma^2 \rmI).
\end{align}
In this case there exists a closed-form expression~\cite{song2020improved, karras2022elucidating} to the optimal denoiser
\begin{align}\label{eq:denoiser}
D(\vx; \sigma) = \frac{\sum_i \mathcal{N}(\vx;\, \vy_i,\, \sigma^2 \mathbf{I}) \, \vy_i}{\sum_i \mathcal{N}(\vx;\, \vy_i,\, \sigma^2 \mathbf{I})}.
\end{align}
In other words, $D(\vx;\, \sigma)$ is a locally weighted mean of data samples $\{\vy_i\}$ around $\vx$ under a Gaussian kernel with bandwidth $\sigma$. The denoising score function can be thought of as a
non-parametric guide on how to update $\vx$ in order to move it towards its \emph{weighted nearest neighbors}.

%% file: 04_app.tex
\section{Score Jacobian Chaining for 3D Generation}\label{sec:app}

Let $\vtheta$ denotes the parameters of a 3D asset, \eg, voxel grid of $ (\mathrm{RGB}, \tau) $ as in Sec.~\ref{sec:voxels}. Our goal is to model and sample from the distribution $p(\vtheta)$ to generate a 3D scene. In our setting, only a pretrained 2D diffusion model on images $p(\vx)$ is given and we do not have access to 3D data. To relate the 2D and 3D distributions $p(\vx)$ and $p(\vtheta)$, we assume that the probability density of 3D asset $\vtheta$ is proportional to the expected probability densities of its multiview 2D image renderings $\vx_\pi$ over camera poses $\pi$, \ie,
\begin{align}\label{eq:p_theta}
p_\sigma(\vtheta) \propto \E_{\pi} \big[ p_\sigma(\vx_\pi(\vtheta)) \big],
\end{align}
up to a normalization constant $Z = \int \E_{\pi} \big[ p_\sigma(\vx_\pi(\vtheta)) \big] d\vtheta.$
That is, a 3D asset $\vtheta$ is as likely as its 2D renderings $\vx_\pi$. 

Next, we establish a lower bound, $\log \tilde{p}_\sigma(\vtheta)$, on the distribution in~\equref{eq:p_theta}  using Jensen's inequality:
\bea
\log p_\sigma(\vtheta) =& \hspace*{-1.45cm}\log \big[ \E_{\pi} (p_\sigma(\vx_\pi)) \big] - \log Z\\
\geq& \E_{\pi} [ \log p_\sigma(\vx_\pi) ] - \log Z \triangleq \log \tilde{p}_\sigma(\vtheta).
\eea
Recall that the score is the \emph{gradient} of log probability density of data. By chain rule
\begin{align}
\grad_{\vtheta} \log\tilde{p}_\sigma(\vtheta) &= \E_{\pi} \left[ \grad_{\vtheta} \log p_\sigma(\vx_\pi) \right ] \\
\pdv{\log\tilde{p}_\sigma(\vtheta)}{\vtheta} &= \E_{\pi} \left[ \pdv{\log p_\sigma(\vx_\pi)}{\vx_\pi}  \cdot \pdv{\vx_\pi}{\vtheta} \right ] \\ 
\label{eq:3D_score}
\underbrace{\grad_{\vtheta}\log\tilde{p}_\sigma(\vtheta)}_{\text{3D score}} 
&= \E_{\pi} [ ~ 
\underbrace{\grad_{\vx_\pi}\log p_\sigma(\vx_\pi)}_{\text{2D score; pretrained}} 
\cdot 
\underbrace{ \mJ_\pi \vphantom{\grad_{\vx_\pi}\log p(\vx_\pi)} }_{\text{renderer Jacobian}} ].
\end{align}
We will next discuss how to compute the 2D score in practice using a pretrained diffusion model.

\input{figures/fig_ood}

\subsection{Computing 2D Score on Non-Noisy Images}
\label{sec:compute_render_score}
Computing the 3D score in~\equref{eq:3D_score} requires the 2D score on $\vx_\pi$. A first attempt would be to directly apply the score from the denoiser in ~\equref{eq:score_approx}, \ie,
\bea
\label{eq:naive_score}
\text{score}(\vx_\pi, \sigma) \triangleq ({D}(\vx_\pi; \sigma) - \vx_\pi)/{\sigma^2}.
\eea
Unfortunately, evaluating the pretrained denoiser $D$ on $\vx_\pi$ causes an out-of-distribution (OOD) problem. 
From the training objective in~\equref{eq:training}, at each noise level 
$\sigma$, the denoiser $D$ has only seen noisy inputs of the distribution $\vy + \sigma \vn$ where $ \vy \sim p_\text{data}$ and $\vn \sim \gauss{0}{\mathbf{I}}$. However, %
a rendered image $\vx_\pi$ from 3D asset $\vtheta$ is generally not consistent with such distribution. 

We illustrate this OOD situation in~\figref{fig:ood}.
Given a denoiser pretrained on FFHQ~\cite{stylegan} by~\citet{baranchuk2021label}, we visualize the output $D(\vx_{\text{blob}}; \sigma=6.5)$ where the input $\vx_{\text{blob}}$ is a non-noisy image showing an orange blob centered on a grey canvas. Under the intuition that $D$ predicts a \emph{weighted nearest neighbor} as reviewed in~\eqref{eq:denoiser}, 
we expect the denoiser to blend the orange blob with the manifold of faces. %
However in reality we observe sharp artifacts when updating with this score $(D(\vx_{\text{blob}}; \sigma) - \vx_{\text{blob}}) / \sigma^2 $ and the image becomes further away from the face manifold. %

\begin{figure}[!tb]
    \centering
    \includegraphics[width=0.85\linewidth]{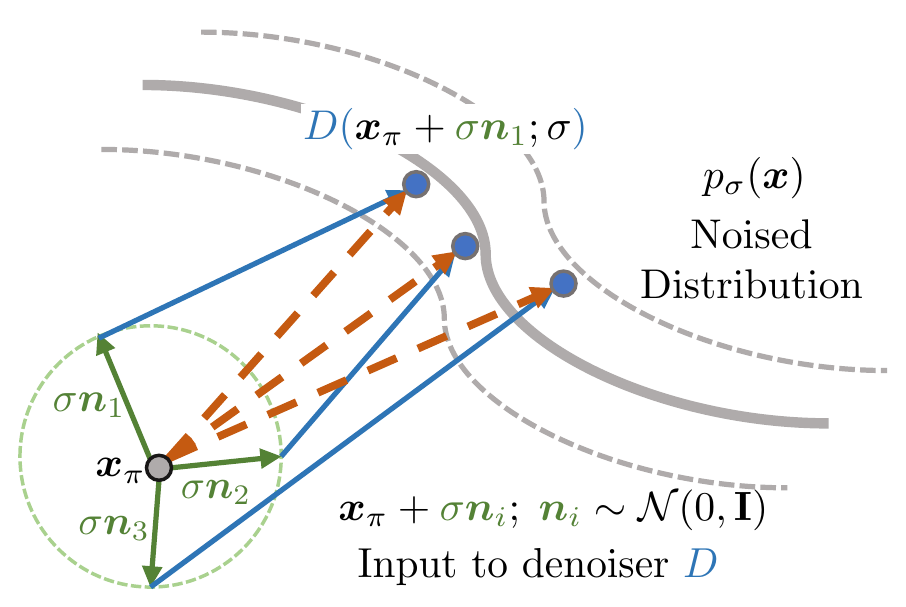}
    \vspace{-0.2cm}
    \caption{
        Computing $\nescore{}$ on 2D renderings $\vx_\pi$. Directly evaluating $D(\vx_\pi; \sigma)$ leads to an OOD problem. Instead, we add noise to $\vx_\pi$, and evaluate $D(\vx_\pi + \sigma \vn; \sigma)$ (\textcolor{Ddotcol}{blue dots}). The $\nescore{}$ is then computed by averaging over the \textcolor{mybrown}{brown dashed} arrows, corresponding to multiple samples of $\vn$.
        See~\secref{sec:compute_render_score} for details. 
        }
    \label{fig:score_x_pi}
    \vspace{-0.25cm}
\end{figure}

\smallskip
{\bf\noindent \escore{}.}
To address the OOD problem, we propose \emph{\escore{}} ($\nescore{}$). It computes the score on non-noisy images $\vx_\pi$ with a denoiser $D$ by adding noise to the input, and then considering the expectation of the predicted scores \wrt the random noise,
\begin{align}
\label{eq:score_algorithm}
&\nescore{}(\vx_\pi, \sqrt{2}\sigma) \\
\triangleq & \E_{\vn \sim \gauss{0}{\rmI}}~ \left[\mathrm{score}(\vx_\pi + \sigma\vn, \sigma)\right] \\ 
=& \E_{\vn}~ \scalemath{0.9}{
    \left[\frac{{D}(\vx_\pi + \sigma\vn, \sigma) - (\vx_\pi + \sigma\vn)}{\sigma^2}\right] 
} \label{eq:before_ens} \\ 
=& \E_{\vn}~ \scalemath{0.9}{ 
    \left[ \frac{{D}(\vx_\pi + \sigma\vn, \sigma) - \vx_\pi}{\sigma^2} \right] 
    \bcancel{
        - \underbrace{\E_{\vn}\left[\frac{\vn}{\sigma}\right]}_{\text{=$0$}}
    }\label{eq:ens}
}.
\end{align}
In practice, we use the Monte Carlo estimate of the expectation in~\cref{eq:ens}.
The algorithm is illustrated in \figref{fig:score_x_pi}. Given a set of sampled noises $\{\vn_i \}$, each $D(\vx_\pi + \sigma \vn_i)$ provides an update direction on the perturbed input $\vx_\pi + \sigma \vn_i$. By averaging over the noise perturbations $\{\vn_i \}$, 
we obtain an update direction on $\vx_\pi$ itself.

\smallskip
{\noindent \bf Justifying $\nescore{}$ in~\cref{eq:score_algorithm}.}
We show that \escore{} provides an approximation to the score on $\vx_\pi$ at 
an inflated noise level of $\sqrt{2}\sigma$
\bea 
\nescore{}(\vx_\pi, \sqrt{2}\sigma) \approx \score{\vx_\pi}{\sqrt{2}\sigma}.
\eea
\vspace{-0.25cm}
\begin{mdframed}[style=MyFrame,align=center]
\begin{lemma}\label{thm:density}
Assuming an empirical data distribution $p_{\sigma}(\vx)$ in~\equref{eq:gmm}, for any $\vx \in \R^d$
\bea\label{eq:density_dilate}
\log p_{\sqrt{2}\sigma}(\vx) \ge \E_{\vn \sim \gauss{0}{\rmI}} \log p_\sigma(\vx + \sigma\vn).
\eea
\end{lemma}
\end{mdframed}
\begin{proof}
Observe that the LHS of~\equref{eq:convolve} is a convolution of two Gaussians, therefore
\bea\label{eq:convolve}
\E_{\vn \sim \gauss{0}{\rmI}}~ [\mathcal{N}(\vx + \sigma \vn;\, \vmu,\, \sigma^2 \rmI) ] 
= \mathcal{N}(\vx;\, \vmu,\, 2\sigma^2 \rmI)
\eea
Recall that $p_{\sigma}(\vx)$ is a mixture of Gaussians per~\equref{eq:gmm};
\begin{align}
\mkern-18mu \scalemath{0.9}{p_{\sqrt{2}\sigma}(\vx)}  &= \E_{\vy \sim p_\text{data}}~ \mathcal{N}(\vx;\, \vy, \, 2\sigma^2 \rmI) \label{eq:left}  \\
&= \E_{\vy \sim p_\text{data}}~ \E_{\vn \sim \gauss{0}{\rmI}}~ \mathcal{N}(\vx+ \sigma\vn;\, \vy, \, \sigma^2 \rmI)  \\
&= \E_{\vn \sim \gauss{0}{\rmI}}~ \E_{\vy \sim p_\text{data}}~ \mathcal{N}(\vx+ \sigma\vn;\, \vy, \, \sigma^2 \rmI) \\
&= \E_{\vn \sim \gauss{0}{\rmI}}~ p_\sigma(\vx + \sigma\vn). \label{eq:right}
\end{align}
Taking the $\log$ on both sides of ~\equref{eq:right} and by Jensen's inequality, we arrive at~\equref{eq:density_dilate}.
\end{proof}

\begin{mdframed}[style=MyFrame,align=center]
\begin{claim}\label{thm:claim1}

Assuming a trained denoiser $D$ as in~\equref{eq:score_approx}, our $\nescore{}(\vx_\pi, \sqrt{2}\sigma)$ in~\cref{eq:score_algorithm} computes the gradient \wrt a lower bound of $\log p_{\sqrt{2}\sigma}(\vx)$.
\end{claim}
\end{mdframed}
\begin{proof}
By taking the gradient of the RHS of~\cref{eq:density_dilate},
\bea\nonumber
\grad_\vx~ \E_{\vn} \log p(\vx + \sigma\vn, \sigma) =& \E_{\vn} \score{\vx + \sigma\vn}{\sigma} \\
=& \hspace*{-0.43cm} \E_{\vn} [\text{score}(\vx_\pi+\sigma\vn, \sigma)].
\eea
which is the proposed $\nescore{}$ algorithm
in~\cref{eq:score_algorithm}. 
\end{proof}

\subsection{Inverse Rendering on Voxel Radiance Field}\label{sec:voxels}
With the computation of the 2D score resolved, the other half of our setup in \equref{eq:3D_score} requires access to the Jacobian of a differentiable renderer. 

\smallskip
{\noindent \bf 3D Representation.} We represent a 3D asset $\vtheta$ as a voxel radiance field~\cite{plenoxels,dvgo,tensorf}, which is much faster to access and update compared to a vanilla NeRF parameterized by a neural network~\cite{nerf}. The parameters $\vtheta$ consist of a density voxel grid $ \mathbf{V}^{(\text{density})} \in \R^{1 \times N_x \times N_y \times N_z} $ and 
a voxel grid of appearance features $ \mathbf{V}^{(\text{app})} \in \R^{c \times N_x \times N_y \times N_z} $. Conventionally the appearance features are simply the $\mathrm{RGB}$ colors
and $c = 3$. For simplicity, we do not model view dependencies in this work. 

\smallskip
{\noindent \bf Inverse Volumetric Rendering.} Image rendering is performed independently
along a camera ray through each pixel. 
We cut a camera light ray into equally distanced segments of length $d$, 
and at the spatial location corresponding to the beginning of the $i$-th segment 
we sample a $(\mathrm{RGB}_i, \tau_i)$ tuple from the color and density grids using trilinear interpolation. These values are alpha-composited using volume rendering quadrature~\cite{maxvol} into the pixel color $C = \sum_i w_i \cdot \mathrm{RGB}_i$, where
\begin{align}
     \label{eq:volrend} 
    w_i = \alpha_i \cdot \prod_{j=0}^{i-1} (1 - \alpha_j);\quad \alpha_i = 1 - \exp(-\tau_i d). 
\end{align}
Volume rendering of $\vtheta$ is directly differentiable. At a rendered image $\vx_\pi$, the Vector-Jacobian product in~\equref{eq:3D_score} between $\nescore({\vx_\pi})$ and the Jacobian $J_\pi = \pdv{\vx_\pi}{\vtheta}$ is computed by back-propagating the score through~\equref{eq:volrend}. This Vector-Jacobian product provides us with the 3D gradient needed for generative modeling on the voxel radiance field.

\input{figures/fig_2dcomp}

\smallskip
{\noindent \bf Regularization Strategies.} 
The voxel grid is a very powerful 3D representation for volumetric rendering. Given noisy 2D guidance, the model may cheat by populating the entire grid with small densities such that the combined effect from one view hallucinates a plausible image. We propose several techniques to encourage the formation of a coherent 3D structure.

\smallskip
{\noindent\it Emptiness Loss:}  Ideally, the space should be sparse with near zero densities except at the object. We propose an emptiness loss to encourage sparsity on a ray $\mathbf{r}$:
\bea\label{eq:emptiness}
\gL_{\text{emptiness}}(\mathbf{r}) = \frac{1}{N} \sum_{i=1}^{N} \log (1+\beta \cdot w_i ),
\eea
where $w_i$ are the alpha-composited weights shown in \eqref{eq:volrend}. 
The shape of the $\log$ function imposes severe penalties at the onset of small weights, but does not grow aggressively if the weights are large. It is consistent with our aim to eliminate
small densities. The hyperparameter $\beta$ controls the steepness of the loss function near 0. A larger $\beta$ will put more emphasis on eliminating low-density noise. We set $\beta = 10$. 

\smallskip
{\noindent\it Emptiness Loss Schedule:} 
We use a hyperparameter $\lambda$ to control the contribution of the emptiness loss. If we apply a large emptiness loss, it will hinder the learning of geometry in the early stage of training. But if the emptiness is too small, there will be floating density artifacts. We adopt a two-stage noise elimination schedule to deal with this problem. In the first $K$ iterations, we use a relatively small weighting factor $\lambda_1$. After the $K^{\text{th}}$ iteration, it is increased to a larger $\lambda_2$. In our experiments $\lambda_1=1\times 10^4$ and $\lambda_2=2\times 10^5$. We provide an ablation study of this technique in~\cref{fig:emptiness_ablations} to show its effectiveness.

\input{figures/fig_qual_results}

\smallskip
{\noindent\it Center Depth Loss:} Sometimes the optimization places the object away from the scene center. The object either becomes small or wander around the image boundary. For the few cases when this happens we apply a center depth loss
\begin{equation}\label{eq:center_depth_loss}
\gL_{\text{center}}(\mD) = -\log\left(\frac{1}{|\mathcal{B}|}\sum_{\mathbf{p}\in \mathcal{B}} \mD(\mathbf{p}) - \frac{1}{|\mathcal{B}^\complement|}\sum_{\mathbf{q}\notin \mathcal{B}} \mD(\mathbf{q})\right)
\end{equation}
where $\mD$ is the depth image, $\mathcal{B}$ is a box (set of pixel locations) %
at the center of the image, and $\mathcal{B}^\complement$ is its complement.

\subsection{SJC vs. DreamFusion}\label{sec:compare_dream_fusion}
In this section, we describe the differences and the connections between our SJC and DreamFusion.

{\bf\noindent Differences from DreamFusion.}
In terms of formulation, DreamFusion's computation of the gradient \wrt $\theta$ involves {a U-Net Jacobian term} (see Eq.~2 in their paper~\cite{dreamfusion}). In practice, they ``found that omitting the U-Net Jacobian term'' to be more effective. On the other hand, this U-Net Jacobian term \textit{does not appear} in our formulation. Their additional justification in the appendix actually leans more towards our viewpoint. 
An additional contribution of ours beyond DreamFusion~\cite{dreamfusion} is our analysis of the effect that the OOD problem has when using a denoiser on rendered images (Claim~\ref{thm:claim1}), and the \nescore method to address it. 
For the variance reduction technique, namely the use of the Monte-Carlo estimate on~\cref{eq:ens}, or $\hat{\bm{\epsilon}} - \bm{\epsilon}$ (in DreamFusion), vs. on~\cref{eq:before_ens}, we observe comparable performance between the two methods empirically for 3D generation.

{\bf\noindent Influences by DreamFusion.}
At the time of this submission, DreamFusion is a concurrent arXiv paper. However, as we have read the paper, our research was influenced by their reported observations. In particular, we adopted the idea of randomized scheduling of $\sigma$ during 3D optimization for easier hyperparameter tuning, and used view-augmented language prompting that improves the overall 3D quality. For future work we do hope to explore a more general solution than view-dependent prompts.

%% file: figures/fig_ood.tex
\begin{figure}[t]
    \centering
    \setlength{\tabcolsep}{3pt}
    \begin{tabular}{ccc}
    Input $\vx_{\text{blob}}$ & $D(\vx_{\text{blob}}, \sigma)$ & $D(\vx_{\text{blob}} + \sigma \vn, \sigma)$\\
    \includegraphics[width=0.31\linewidth]{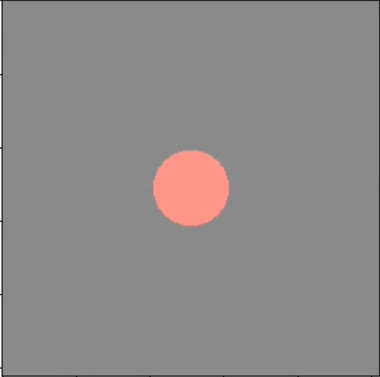} &
    \includegraphics[width=0.31\linewidth]{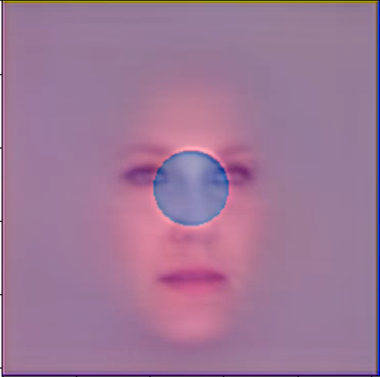} & 
    \includegraphics[width=0.31\linewidth]{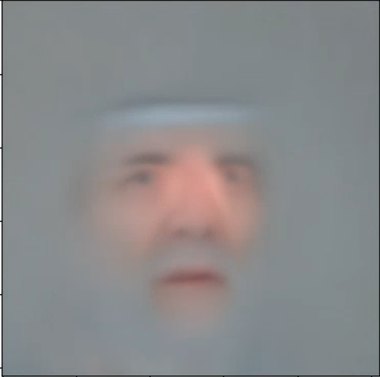}
    \end{tabular}
    \vspace{-0.3cm}
    \caption{Illustration of denoiser's OOD issue using a denoiser pretrained on FFHQ. 
    When directly evaluating $D(\vx_{\text{blob}}, \sigma)$ the model did not correct for the orange blob into a face image. Contrarily, evaluating the denoiser on noised input $D(\vx_{\text{blob}} + \sigma \vn, \sigma)$ produces an image that successfully merges the blob with the face manifold.
    }
    \label{fig:ood}
\end{figure}

%% file: figures/fig_2dcomp.tex
\begin{figure}[t]
    \centering
    \setlength{\tabcolsep}{2pt}
    \renewcommand{\arraystretch}{1.3}
    \begin{tabular}{ccccc}
    & \multicolumn{2}{c}{Annealed $\sigma$} & \multicolumn{2}{c}{Random $\sigma$}\\
    \rotatebox{90}{\hspace{0.33cm} FFHQ}  & \includegraphics[width=0.22\linewidth]{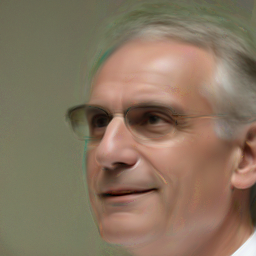} & 
    \includegraphics[width=0.22\linewidth] {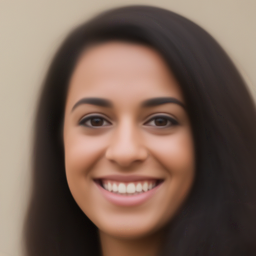} &
    \includegraphics[width=0.22\linewidth]{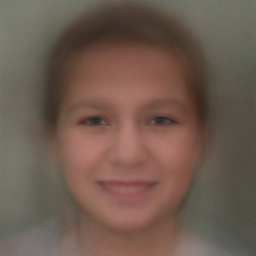} &
    \includegraphics[width=0.22\linewidth]{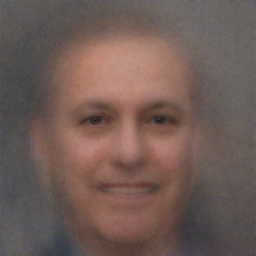}\\
    \rotatebox{90}{\hspace{0.33cm} LSUN} & 
    \includegraphics[width=0.22\linewidth]
    {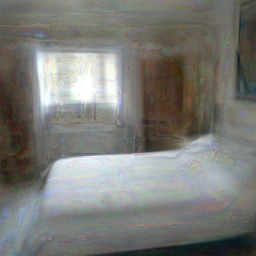} &
    \includegraphics[width=0.22\linewidth]{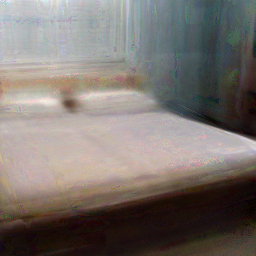} &
    \includegraphics[width=0.22\linewidth]{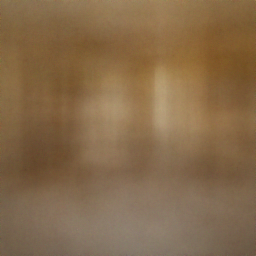} &
    \includegraphics[width=0.22\linewidth]{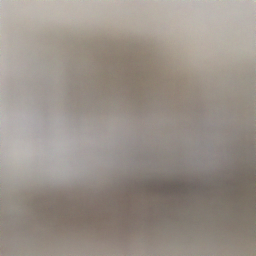}\\
    \rotatebox{90}{\;SD, scale=3} & 
    \includegraphics[width=0.22\linewidth]{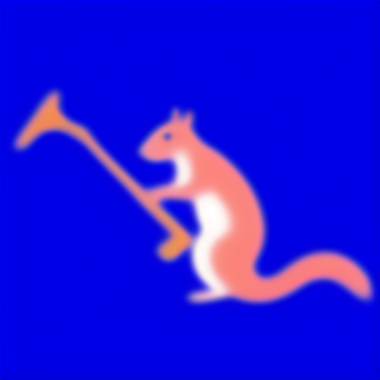} &
    \includegraphics[width=0.22\linewidth]{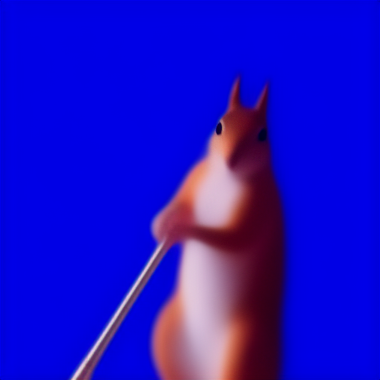} &
    \includegraphics[width=0.22\linewidth] {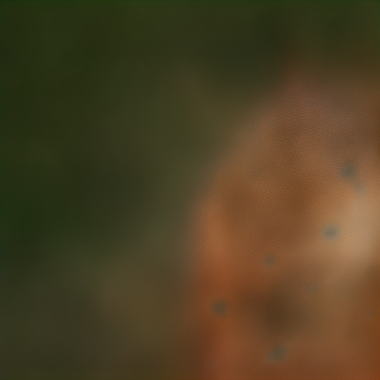} &
    \includegraphics[width=0.22\linewidth]{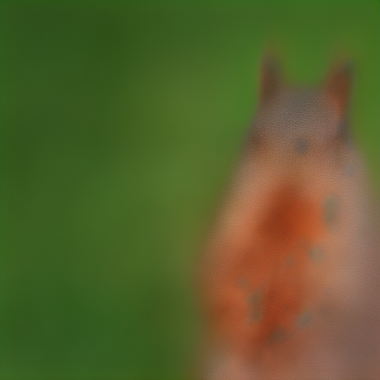} \\
    \rotatebox{90}{SD, scale=10}  &
    \includegraphics[width=0.22\linewidth]{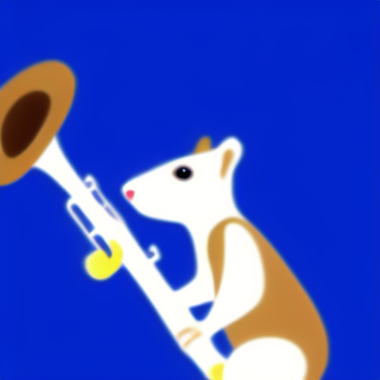} &
    \includegraphics[width=0.22\linewidth]{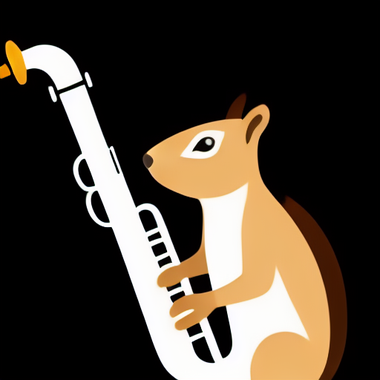} &
    \includegraphics[width=0.22\linewidth]{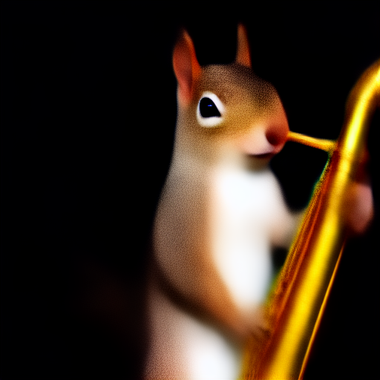} &
    \includegraphics[width=0.22\linewidth]{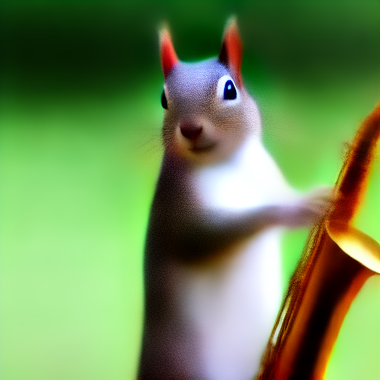} \\
    \end{tabular}
    \vspace{-0.25cm}
    \caption{
        Sampling 2D images with \escore{}. We compare Annealed vs Random $\sigma$
        schedule against several diffusion models. Row 1 \& 2: the random $\sigma$ schedule 
        exhibits strong mode-seeking behavior, and it results in low-quality ``mean'' images
        on unconditioned diffusion models trained on FFHQ and LSUN Bedroom. In this case,
        we need a carefully designed annealed $\sigma$ schedule to produce better, more diverse samples. Row 3 \& 4: Stable Diffusion (SD) is conditioned on the prompt ``a squirrel holding a saxophone''. The use of natural language makes the conditioned distribution 
        much easier to sample from. When the guidance scale is elevated to 10, Random
        $\sigma$ schedule that fails on FFHQ and LSUN starts to produce crisp, clean images. 
    }
    \vspace{-0.5cm}
    \label{fig:2dcomp}
\end{figure}

%% file: figures/fig_qual_results.tex
\begin{figure*}[!tbh]
    \centering
    \setlength{\tabcolsep}{2pt}
    \renewcommand{\arraystretch}{0.95}
    \begin{tabular}{cc}
    \small A DSLR photo of a yellow duck & \small A ficus planted in a pot\\
    \includegraphics[width=0.49\linewidth]{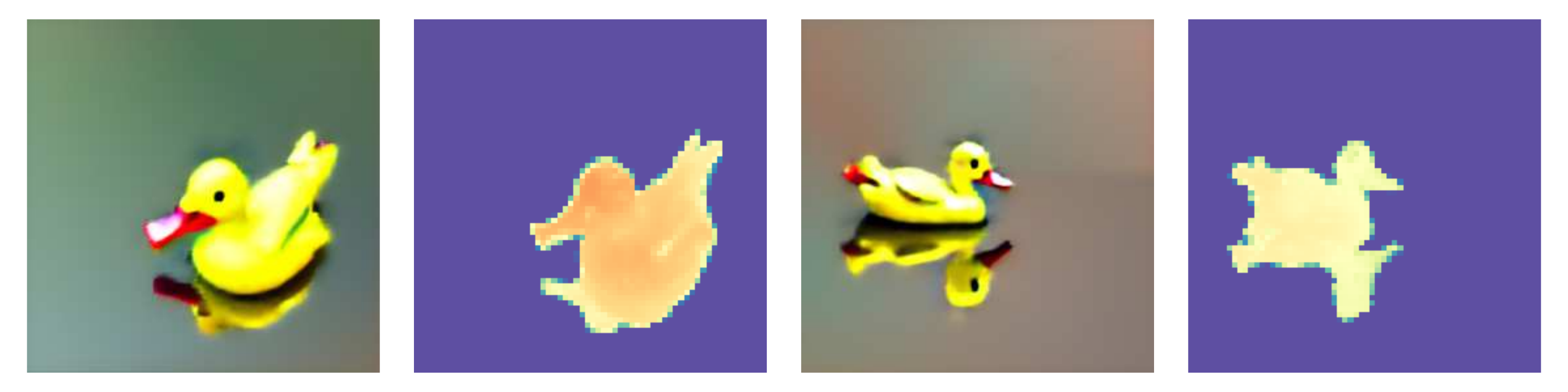} & 
    \includegraphics[width=0.49\linewidth]{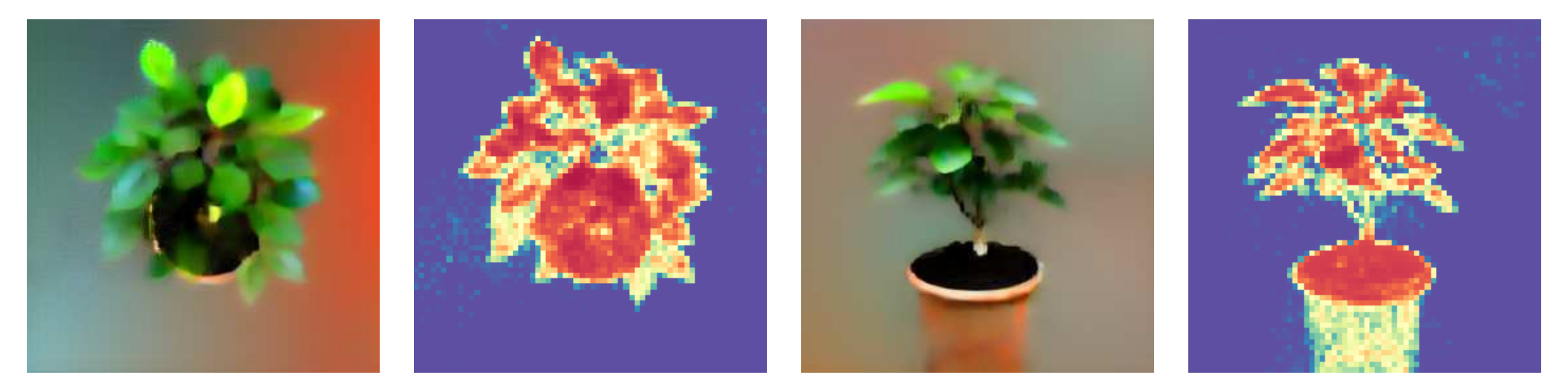}\\
    \small A zoomed out photo a small castle & \small A high quality photo of a toy motorcycle\\
    \includegraphics[width=0.49\linewidth]{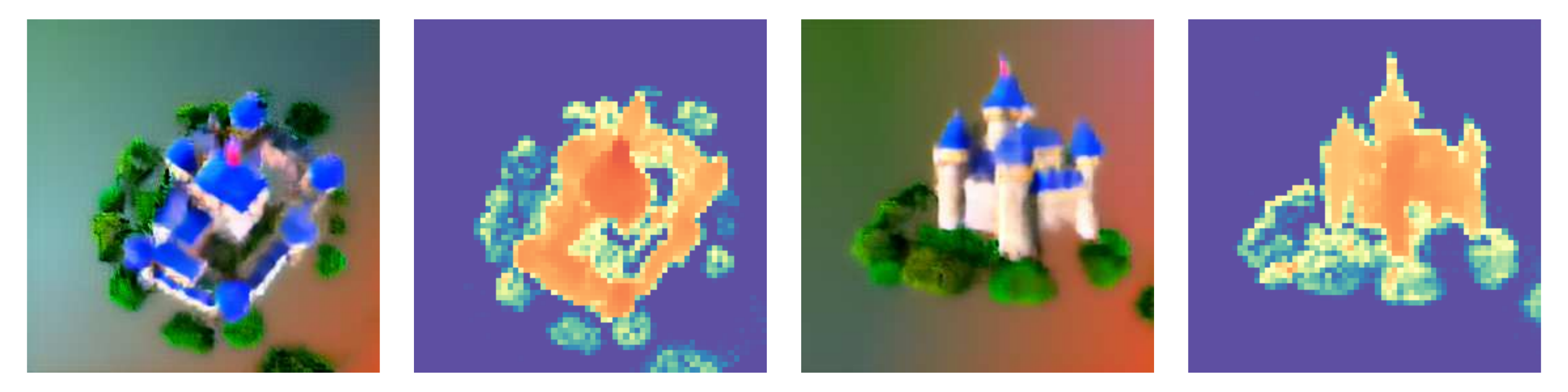} &  
    \includegraphics[width=0.49\linewidth]{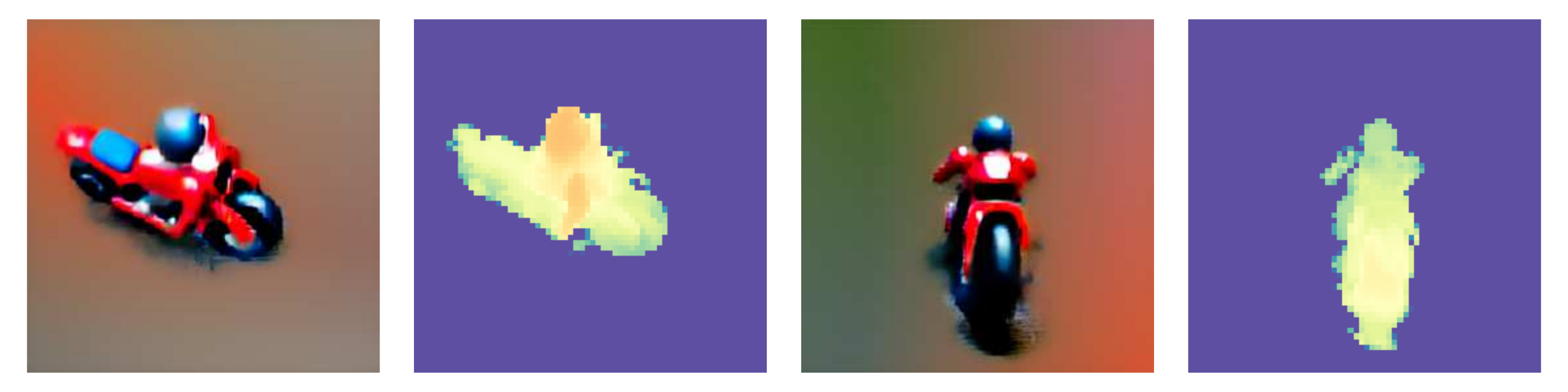}\\
    \small A zoomed out high quality photo of Sydney Opera House & A photo of a horse walking \\
    \includegraphics[width=0.49\linewidth] {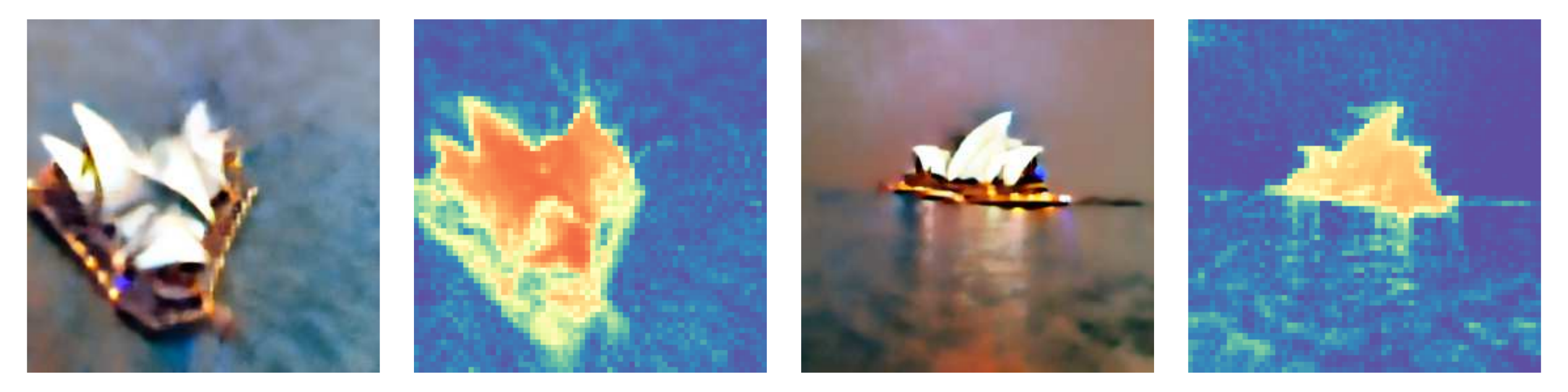} & 
    \includegraphics[width=0.49\linewidth] {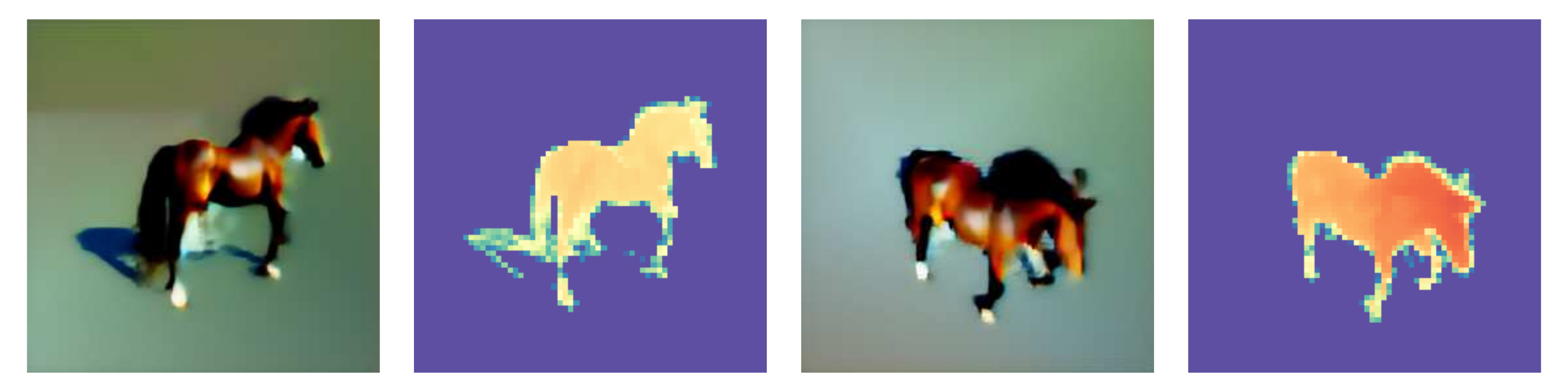} \\
    \end{tabular}
    \vspace{-0.4cm}
    \caption{Qualitative results of text-prompted generation of 3D models with SJC, purely from the pretrained Stable Diffusion (2D) image model. Each row shows two views, with associated depth maps (blue is far, red is near), for a single 3D model generated for a given prompt. Note the detailed appearance as well as a sharp, well-defined depth structure. %
    }
    \vspace{-0.25cm}
    \label{fig:qual_results}
\end{figure*}

%% file: 05_exp.tex
\input{figures/comparison_larger}

\section{Experiments}
We conduct experiments on both unconditioned and conditioned diffusion models to have a more comprehensive understanding of the properties of SJC. %

\smallskip
{\noindent \bf DDPMs trained on FFHQ and LSUN Bedroom} are unconditioned diffusion 
models with an architecture based on the implementation by~\citet{dbeatgan}. 
They are trained on an image resolution of $256 \times 256$. 
FFHQ~\cite{stylegan} is a dataset of aligned faces with diverse coverage of gender, age, race, facial appearance as well as head poses. LSUN Bedroom~\cite{lsun} includes bedroom images with varied furniture layout plans and rich interior design styles.

\smallskip
{\noindent \bf Stable Diffusion} is an expanded work based on Latent Diffusion Model (LDM) developed by~\citet{rombach2022high}. It is trained on the
 LAION5B dataset~\cite{laion}. We use the release version v1.5. 
Diffusion is performed on a latent space of $4 \times 64 \times 64$, then upsampled to $3 \times 256 \times 256$ by a decoder. The model is natively trained for text-conditioned image generations, and exposes a guidance scale parameter that controls the strength of language conditioning~\cite{ho_guidance}. Intuitively a larger guidance scale makes the conditioned image distribution more faithful to the text prompt by trading off sample diversity.

\subsection{Validating $\nescore{}$ on 2D images.}
Before directly jumping to 3D generation, we first verify that $\nescore{}$ provides
effective guidance on a simple 2D image canvas. In other words, here $\vtheta$ is a grid of RGB values and $f$ is an identity function. The hope is that gradient descent on the vector
field produced by $\nescore$ creates high quality images. Here an important decision to make is the schedule of $\{\sigma_i\}$ at which we compute $\nescore{}$. 

\input{figures/emptiness_ablations}

We experimented with an annealed schedule (Annealed $\sigma$) vs. a random schedule (Random $\sigma$) as proposed in DreamFusion. Under the Annealed $\sigma$ schedule, we start from a large 
$\sigma$ and gradually decrease it as we update the image canvas $\vx$. $\nescore{}$ computed at larger $\sigma$ level attends to high level image structure while smaller $\sigma$ provides
stronger guidance on detailed features. The Random $\sigma$ schedule on the contrary uniformly 
samples a $\sigma$ at every step. We show %
qualitative comparisons 
in~\cref{fig:2dcomp}. 

For unconditioned diffusion models trained on FFHQ, 
we observe that Annealed $\sigma$ performs better
than Random $\sigma$, and the image samples have better pose variation and quality. Particularly, the randomized $\sigma$ exhibits severe mode-seeking behavior converging to average faces. In the case of LSUN Bedroom, the mode-seeking behavior results in a blurry image 
canvas with no content.

On the other hand, natural language prompting plays a critical role when sampling images
with Stable Diffusion. When the language guidance is set to a regular level of $3.0$, the observations are broadly consistent with sampling on FFHQ and LSUN. Random $\sigma$
schedule produces blurry outputs. However, when the guidance scale is elevated to $10.0$, 
Random $\sigma$ schedule begins to generate crisp, clean images and outperforms Annealed $\sigma$
schedule. Despite various sophisticated strategies on Annealed $\sigma$ scheduling (see our code for details), at a high language guidance scale Randomized $\sigma$ remains the better option. We hypothesize that stronger language guidance forces the image distribution to be narrower and more beneficial 
for a mode-seeking algorithm. We acknowledge none of the images in \figref{fig:2dcomp}
can match the sample quality of a standard diffusion inference pipeline, and the right 
way to apply $\nescore$ as gradient for optimization remains an open problem.

\smallskip
\subsection{3D Generation}\label{sec:3dgen}
In this paper, we focus on 3D Generation with the language-conditioned Stable Diffusion model. 
We found that tuning the Annealed $\sigma$ schedule on FFHQ and LSUN Bedroom in 3D domain is difficult
in practice, and leave it as future work. Based on the insights from 2D experiments earlier, 
we use Random $\sigma$ schedule coupled with a high language guidance scale. 

\smallskip
{\noindent \bf Rendering with Latent 3D Features.}
Stable Diffusion economizes compute by performing diffusion modeling on the latent features of a
pretrained AutoEncoder. We therefore choose to render a feature image in this latent space
from a features field~\cite{giraffe,eg3d} represented by a voxel grid in $\R^{4 \times N_x \times N_y \times N_z}$. 

\smallskip
{\noindent \bf Qualitative Comparison.}
In~\cref{fig:qual_results}, we show text-prompted 3D generation results from SJC. It is capable of generating complex 3D models over a diverse set of prompts ranging from animals to the Sydney Opera House. Next, we compare SJC with Stable-DreamFusion, the third-party implementation based on the same pretrained Stable Diffusion model. In~\figref{fig:comparison}, we show qualitative comparisons of generated 3D assets given the same prompt. We observe that SJC generates 3D models with better image quality and more sensible structure than Stable-DreamFusion in a significant number of cases. We acknowledge that both systems exhibit quality fluctutations over different trials, and the point of this comparison is to show that our overall pipeline is competitive.

\smallskip
{\noindent \bf Ablations.}
In~\cref{fig:emptiness_ablations}, we conduct ablations to demonstrate the importance of the  proposed emptiness loss and scheduling of its weight $\lambda$ discussed in~\secref{sec:voxels}. We show results without the emptiness loss, with constant weight $\lambda$ vs our proposed scheduling of $\lambda$. We observe that our complete method (Ours) improves the quality of generated 3D models,~\eg, fewer floating artifacts and better geometry.

%% file: figures/comparison_larger.tex
\begin{figure*}[!tbh]
\centering
\setlength{\tabcolsep}{0pt}
\begin{tabular}{>{\begin{sideways}}c<{\end{sideways}}
@{\hspace{0.3em}}cc@{\hspace{0.3em}}cc@{\hspace{0.3em}}cc}

 & \multicolumn{2}{c}{ \small 
    \begin{tabular}{c}
    (a)
  \end{tabular}} &
\multicolumn{2}{c}{ \small 
    \begin{tabular}{c}
    (b)
  \end{tabular}}&
 \multicolumn{2}{c}{ \small 
    \begin{tabular}{c}
    (c)
  \end{tabular}} \\
  
 {\small~~~~~~~~~~~~Ours} &
 \includegraphics[width = .16\textwidth]{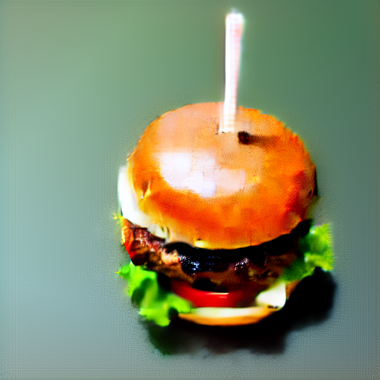}&
 \includegraphics[width = .16\textwidth]{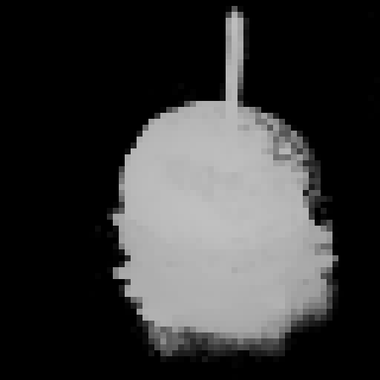}&
 \includegraphics[width = .16\textwidth]{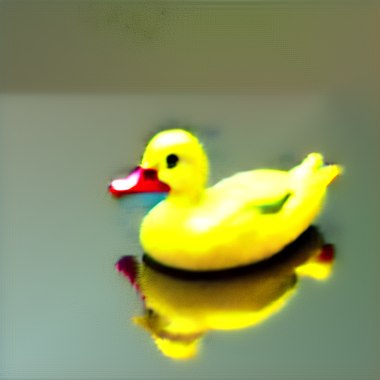}&
 \includegraphics[width = .16\textwidth]{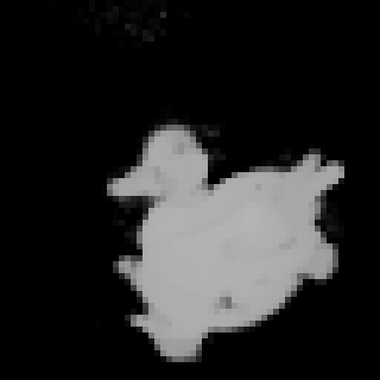}&
 \includegraphics[width = .16\textwidth]{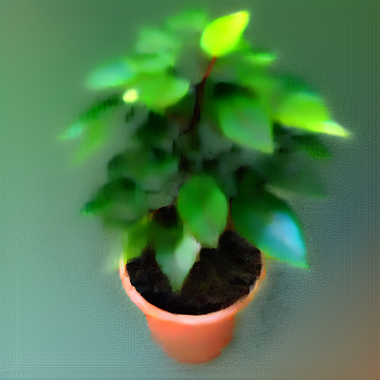}&
 \includegraphics[width = .16\textwidth]{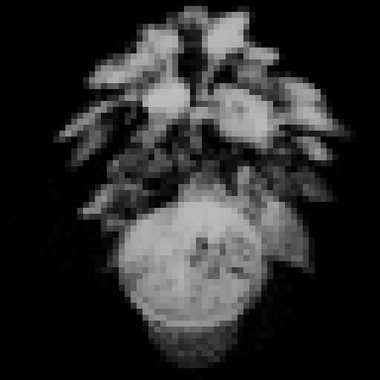}\\

  {\small~~~~~~~~~StableDF} &
 \includegraphics[width = .16\textwidth]{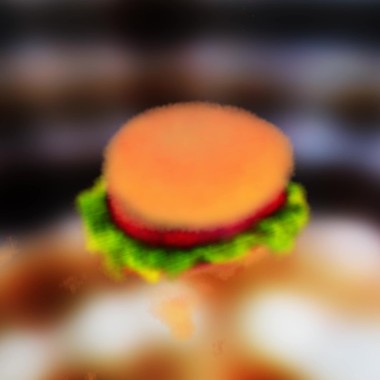}&
 \includegraphics[width = .16\textwidth]{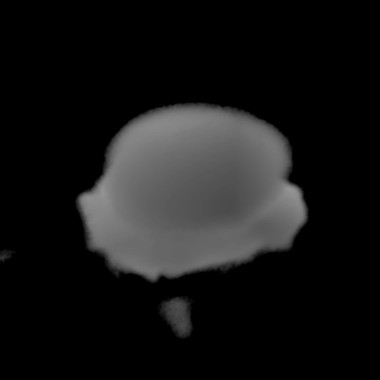}&
 \includegraphics[width = .16\textwidth]{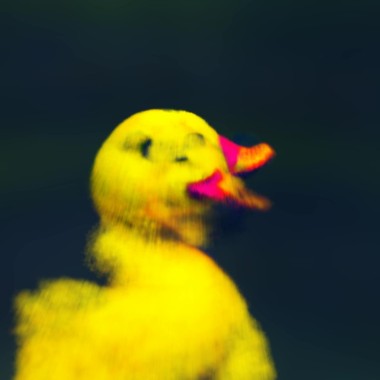}&
 \includegraphics[width = .16\textwidth]{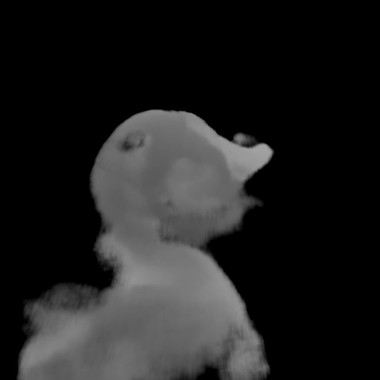}&
  \includegraphics[width = .16\textwidth]{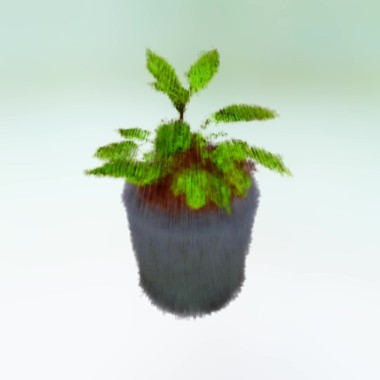}&
 \includegraphics[width = .16\textwidth]{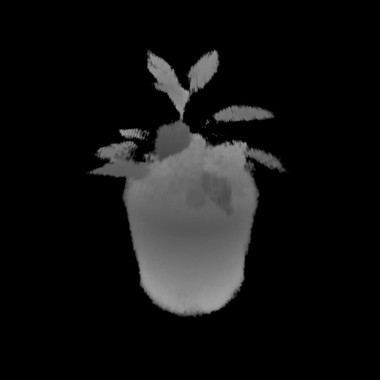}\\

  & \multicolumn{2}{c}{ \small 
    \begin{tabular}{c}
    (d)
  \end{tabular}} &
\multicolumn{2}{c}{ \small 
    \begin{tabular}{c}
    (e)
  \end{tabular}}&
 \multicolumn{2}{c}{ \small 
    \begin{tabular}{c}
    (f)
  \end{tabular}} \\
  
 {\small~~~~~~~~~~~~Ours}&
 \includegraphics[width = .16\textwidth]{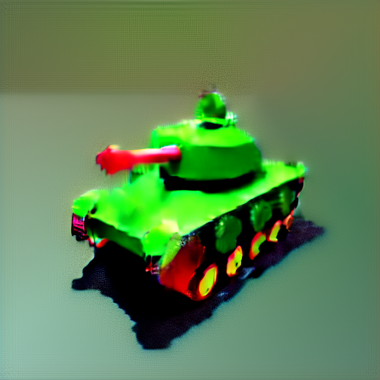}&
 \includegraphics[width = .16\textwidth]{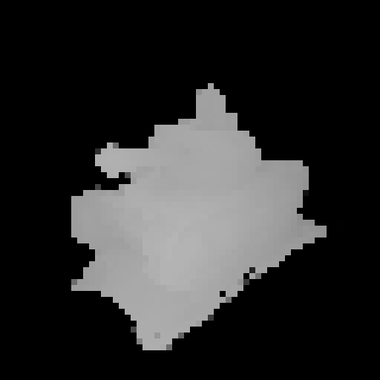}&
  \includegraphics[width = .16\textwidth]{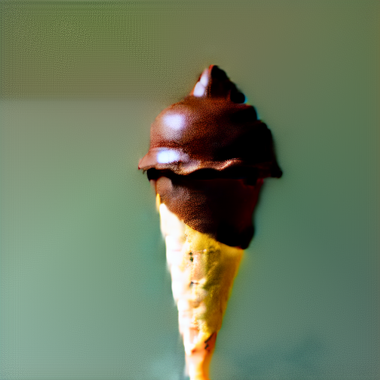}&
 \includegraphics[width = .16\textwidth]{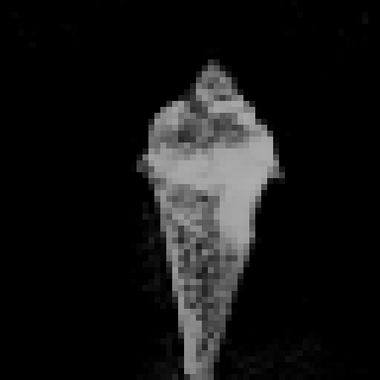}&
  \includegraphics[width = .16\textwidth]{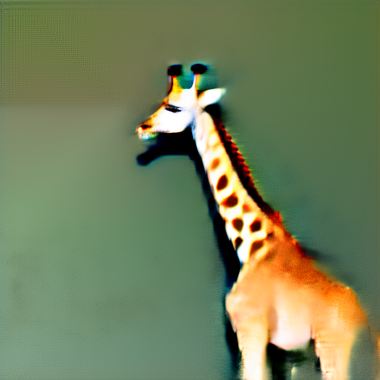}&
 \includegraphics[width = .16\textwidth]{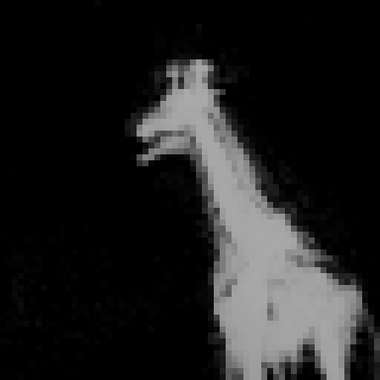}\\

 {\small~~~~~~~~~StableDF} &
 \includegraphics[width = .16\textwidth]{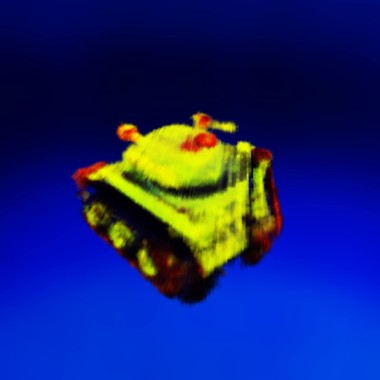}&
 \includegraphics[width = .16\textwidth]{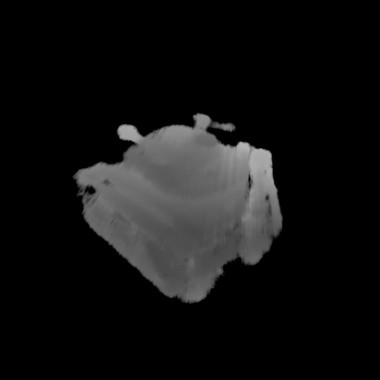}&
 \includegraphics[width = .16\textwidth]{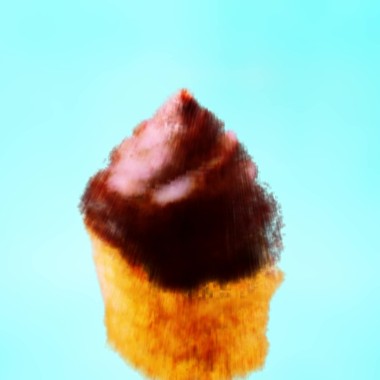}&
 \includegraphics[width = .16\textwidth]{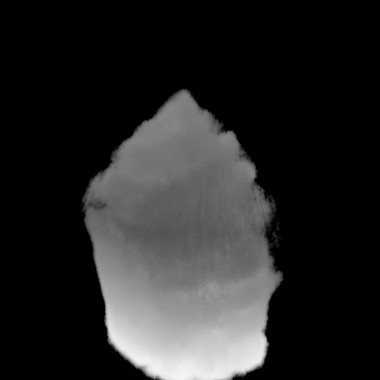}&
 \includegraphics[width = .16\textwidth]{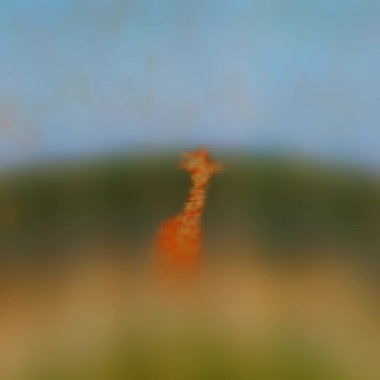}&
 \includegraphics[width = .16\textwidth]{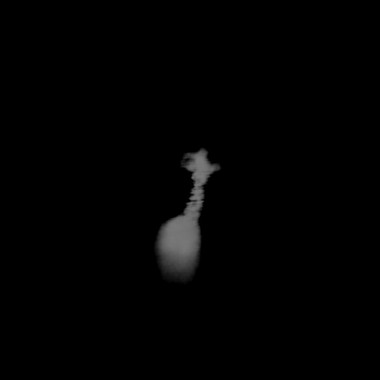}\\

\end{tabular}
\vspace{-0.2cm}
\caption{Qualitative comparison between Stable-DreamFusion (StableDF) and Ours. The prompts are: (a) ``A high quality photo of a delicious burger"; (b) ``a DSLR photo of a yellow duck"; (c) ``A ficus planted in a pot"; (d) ``A product photo of a toy tank"; (e) ``A high quality photo of a chocolate icecream cone"; (f)``A wide angle zoomed out photo of a giraffe". Both methods are run for 10k iterations without per-prompt finetuning on the hyperparameters. The images on the left are rendered RGB images and the images on the right are depth visualization.
}
\vspace{-0.3cm}
\label{fig:comparison}
\end{figure*}

%% file: figures/emptiness_ablations.tex
\begin{figure*}[tbh!]
\centering
\setlength{\tabcolsep}{0pt}
\begin{tabular}{
cc@{\hspace{0.3em}}cc@{\hspace{0.3em}}cc@{\hspace{0.3em}}cc}

 \multicolumn{8}{c}{ \small 
    \begin{tabular}{c}
    A zoomed out high quality photo of Temple of Heaven.
  \end{tabular}
  } \\
 \includegraphics[width = .12\textwidth]{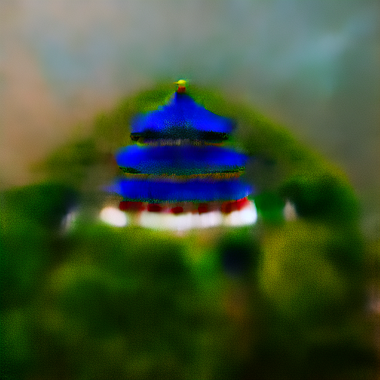}&
 \includegraphics[width = .12\textwidth]{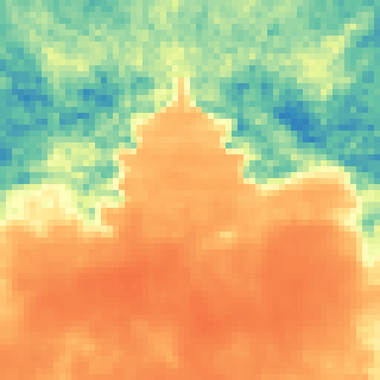}&
 \includegraphics[width = .12\textwidth]{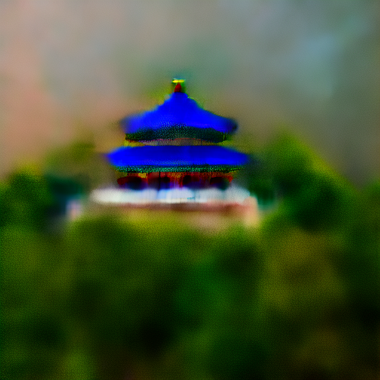}&
 \includegraphics[width = .12\textwidth]{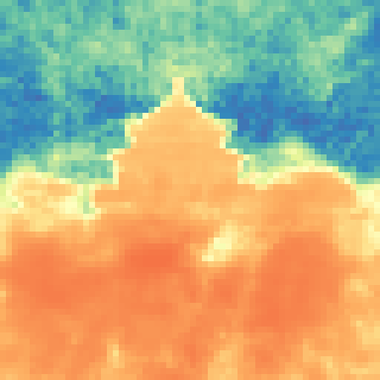}&
 \includegraphics[width = .12\textwidth]{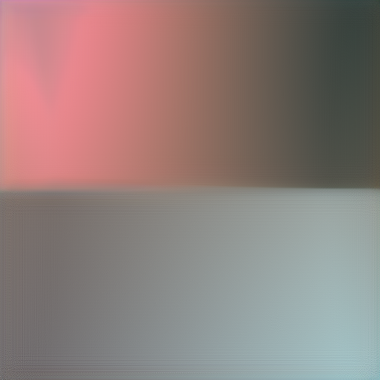}&
 \includegraphics[width = .12\textwidth]{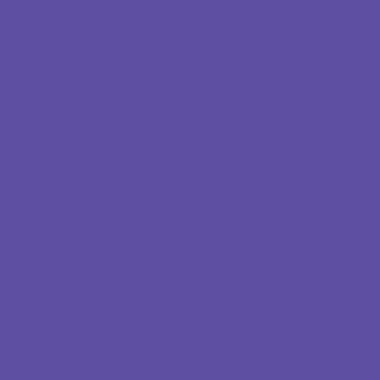}&
 \includegraphics[width = .12\textwidth]{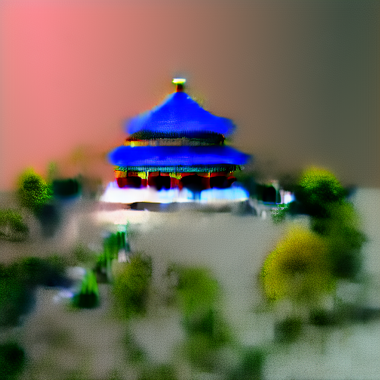}&
 \includegraphics[width = .12\textwidth]{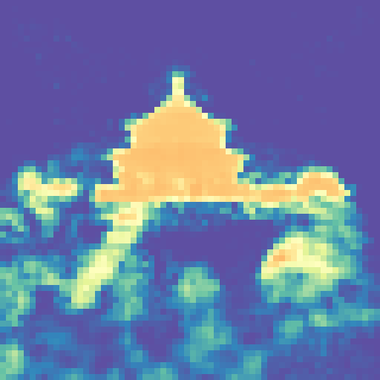}\\
 
 \multicolumn{8}{c}{ \small 
    \begin{tabular}{c}
    a DSLR photo of a rose
  \end{tabular}
  } \\

 \includegraphics[width = .12\textwidth]{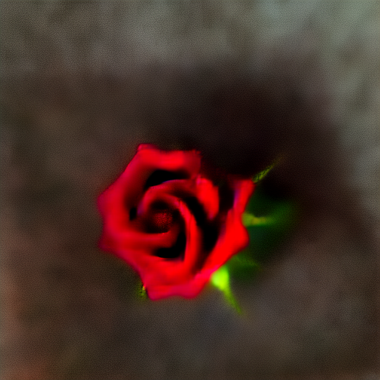}&
 \includegraphics[width = .12\textwidth]{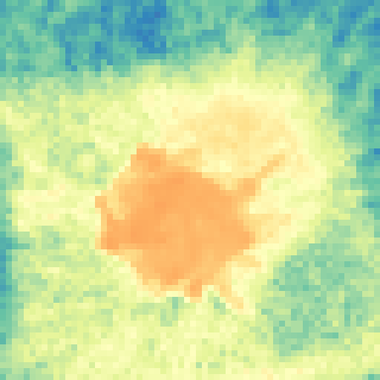}&
 \includegraphics[width = .12\textwidth]{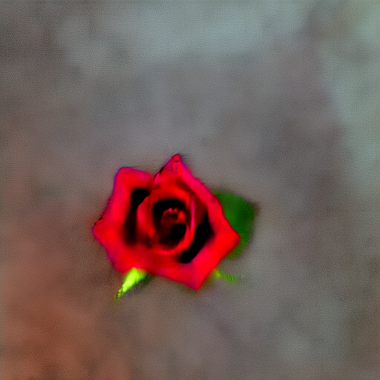}&
 \includegraphics[width = .12\textwidth]{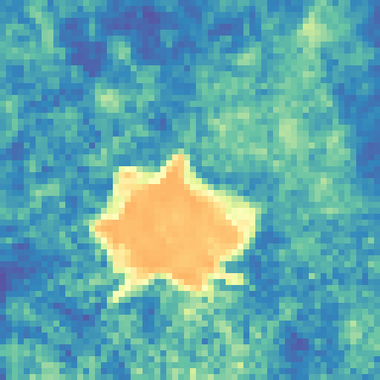}&
 \includegraphics[width = .12\textwidth]{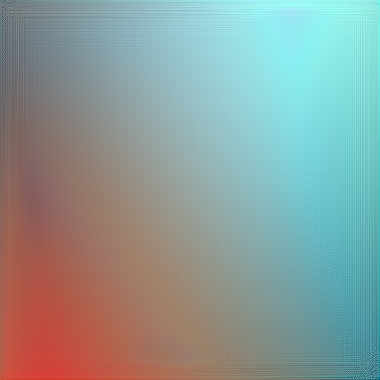}&
 \includegraphics[width = .12\textwidth]{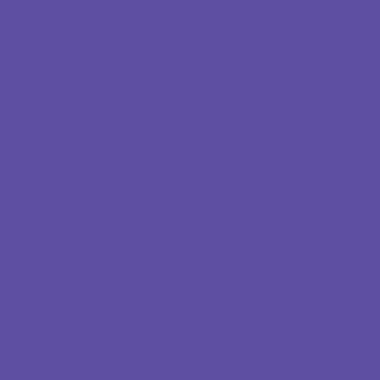}&
 \includegraphics[width = .12\textwidth]{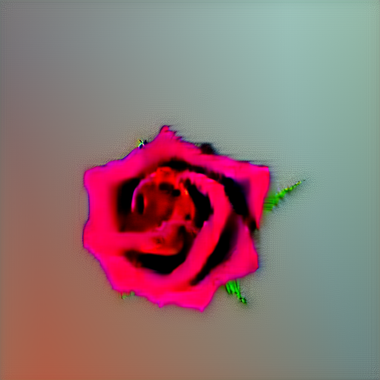}&
 \includegraphics[width = .12\textwidth]{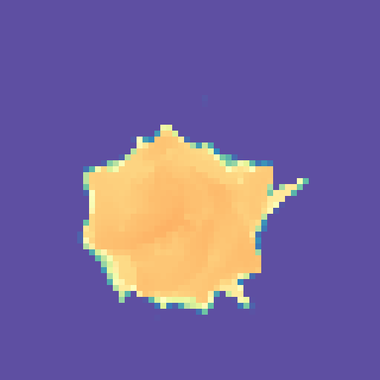}\\
 
 \multicolumn{8}{c}{ \small 
    \begin{tabular}{c}
    A modern house with flat roof floating on water.
  \end{tabular}
  } \\

 \includegraphics[width = .12\textwidth]{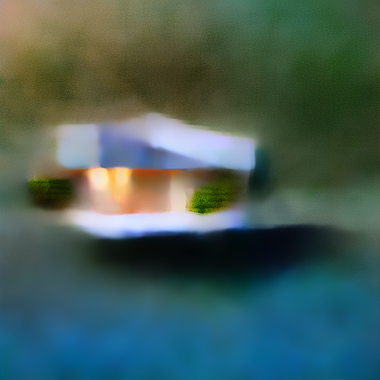}&
 \includegraphics[width = .12\textwidth]{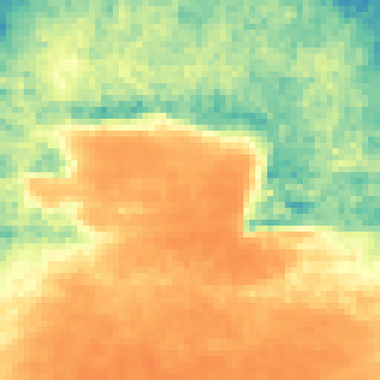}&
 \includegraphics[width = .12\textwidth]{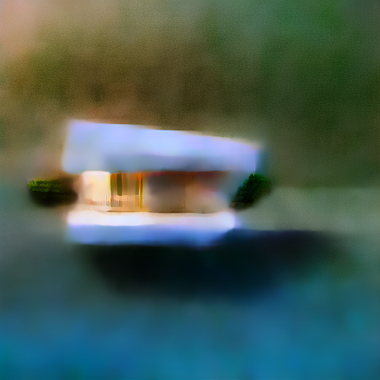}&
 \includegraphics[width = .12\textwidth]{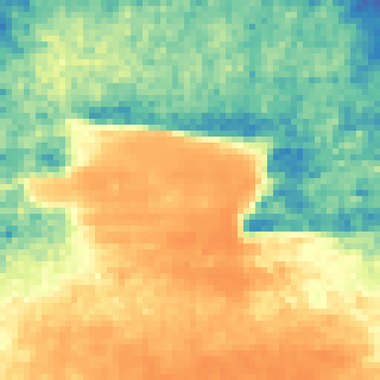}&
 \includegraphics[width = .12\textwidth]{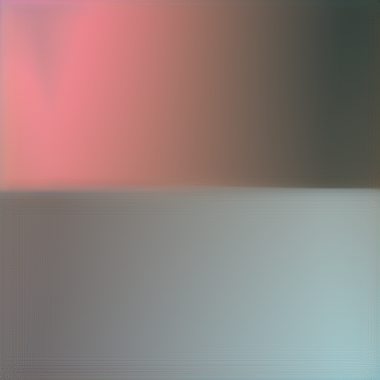}&
 \includegraphics[width = .12\textwidth]{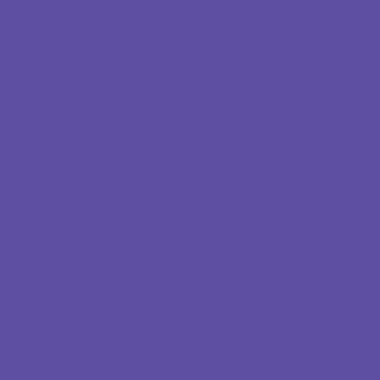}&
 \includegraphics[width = .12\textwidth]{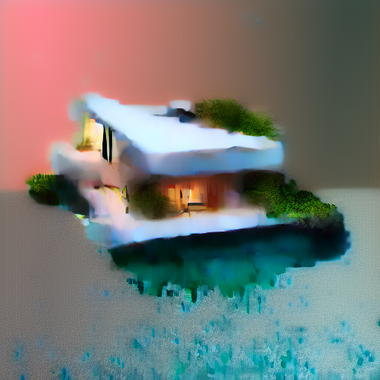}&
 \includegraphics[width = .12\textwidth]{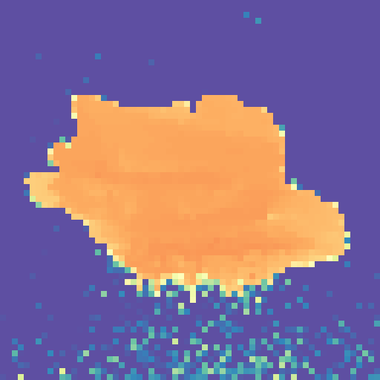}\\

 \multicolumn{2}{c}{ \small 
    \begin{tabular}{c}
    No emptiness loss $(\lambda = 0)$
  \end{tabular}
  } & 
  \multicolumn{2}{c}{ \small 
    \begin{tabular}{c}
    $\lambda = 1\mathrm{e}{4}$
  \end{tabular}
  } &
  \multicolumn{2}{c}{ \small 
    \begin{tabular}{c}
    $\lambda = 2\mathrm{e}{5}$
  \end{tabular}
  } &
  \multicolumn{2}{c}{ \small 
    \begin{tabular}{c}
    {Ours}
  \end{tabular}
  } \\
\end{tabular}
\caption{Ablation experiments on the proposed emptiness loss schedule. For each setting of the loss weight $\lambda$, we show a rendered image and the associated depth map from a randomly sampled viewpoint. Ours incorporates the loss with weight schedule described in~\ref{sec:3dgen}. It leads to better 3D shape, as evidenced in the cleaner depth maps. Setting the loss weight too low yields "cloudy" depth fields. When setting the weight too high, SJC fails to produce meaningful 3D models.} %
\label{fig:emptiness_ablations}
\end{figure*}

%% file: 06_conc.tex
\section{Conclusion}
We propose an optimization-based approach to generate 3D assets from pretrained image (2D) diffusion models. The key technical contribution is the derivation of \escore{} method which bridges the gap between the denoising-trained diffusion models and the non-noisy images encountered in the process of optimizing a 3D model guided by the diffusion. We also propose a new regularization loss for improving the quality of generated 3D scene. Working with the large-scale Stable Diffusion model, we demonstrate that our approach can generate compelling 3D models, comparing favorably to available concurrent work. Finally, we investigate an interesting distinction between the  effect of noise scheduling regime in unconditional image diffusion models and a text-conditional model, and identify an avenue for future work.

\section{Acknowledgements}
The authors would like to thank David McAllester for feedbacks on an early pitch of the work, Shashank Srivastava and Madhur Tulsiani for discussing the $\sqrt{2}$ factor on synthetic experiments. We would like to thank friends at TRI
and 3DL lab at UChicago for suggestions on the manuscript. HC would like to thank Kavya Ravichandran for incredible officemate support, and Michael Maire for the discussion and encouragement while riding Metra.

%% file: 07_appendix.tex
\onecolumn

\appendix

\newcommand{\beginsupplementary}{%
    \setcounter{section}{0}
	\renewcommand{\thesection}{A\arabic{section}}
	\renewcommand{\thesubsection}{\thesection.\arabic{subsection}}

	\renewcommand{\thetable}{A\arabic{table}}%
	\setcounter{table}{0}

	\renewcommand{\thefigure}{A\arabic{figure}}%
	\setcounter{figure}{0}
	
	\setcounter{algorithm}{0}
}
\beginsupplementary

{\noindent\bf \LARGE Appendix %
}
\vspace{0.1cm}

\begin{itemize}
    \item In~\secref{sec:score_perspective}, we provide diffusion models from a score-based perspective following~\citet{karras2022elucidating}.
    \item In~\secref{sec:more_results}, we provide additional experiments on our approach, including additional ablation study, qualitative results, and video results.
    \item In~\secref{sec:impl_details}, we document implementation details. %
\end{itemize}

\input{07_supp_score_based.tex}

\input{07_supp_results.tex}

\input{07_supp_impl.tex}

%% file: 07_supp_score_based.tex
\algrenewcommand\algorithmicindent{0.5em}%
\begin{figure}[h]
\begin{minipage}[t]{0.495\textwidth}
\begin{algorithm}[H]
  \caption{Training} \label{alg:training}
  \small
  \begin{algorithmic}[1]
    \Repeat
      \State $\vx \sim p_{\text{data}}$
      \State $\sigma \sim [\sigma_{\text{min}}, \sigma_{\text{max}}] $
      \State $\vz \sim \gauss{0}{\mathbf{I}} $
      \State Take gradient descent step on
      \Statex $\qquad \grad_\phi \left\| D_\phi(\vx + \sigma \vz,\, \sigma) - \vx  \right\|^2$
    \Until{converged}
    \State $\mathrm{score}(\vx, \sigma) = \score{\vx}{\sigma} = (D_{\phi}(x, \sigma) - x) / \sigma^2$
  \end{algorithmic}
\end{algorithm}
\end{minipage}
\hfill
\begin{minipage}[t]{0.495\textwidth}
\begin{algorithm}[H]
  \caption{Deterministic Sampling} %
  \small
  \begin{algorithmic}[1]
    \State $ \{\sigma_i\}_{i=1}^T $ descending; $\sigma_0 = 0$
    \State $ \vx_T = \sigma_T \vz,~ \vz  \sim \gauss{0}{\rmI}  $
    \For{$t=T, \dotsc, 1$} 
      \State $ \vx_{t-1} = \vx_t + (\sigma_{t} - \sigma_{t-1}) \cdot \sigma_t \cdot \mathrm{score}(\vx_t, \sigma_t) $
      \State  $ \hphantom{\vx_{t-1}} = \underbrace{ (1 - w_t)\, \vx_t + w_t D_{\phi}(\vx_t,\, \sigma_t) }_{\text{weighted average}} \quad w_t = \frac{\sigma_{t} - \sigma_{t-1}}{\sigma_{t}}$
    \EndFor
    \State \textbf{return} $\vx_0$
    \vspace{.04in}
  \end{algorithmic}
\end{algorithm}
\end{minipage}
\caption{
Training and Sampling Algorithm Card for Score-Based Methods with numerical scaling $s(t) = 1$ and $\sigma(t) = t$. Note that the inference step is analogous to DDIM~\cite{ddim}, and simplifies to a weighted averaging between the current iterate $\vx_t$ and the denoiser output $D(\vx_t, \sigma_t)$. This particular
scheduling allows for taking large step sizes, and a sample can be generated in as few as 80 network evaluations~\cite{karras2022elucidating} while maintaining high image quality. 
}
\label{fig:score_algo}
\end{figure}

\section{Diffusion Models from Score-Based Perspective}\label{sec:score_perspective}
We provide a more detailed recap of diffusion models from the score-based perspective. 
For a quick overview, we summarize the training and deterministic sampling
algorithms in~\figref{fig:score_algo}; the deterministic sampling algorithm can be 
made stochastic by adding noise and adjusting $\sigma$ level after each update (details see~\citet{karras2022elucidating}). 

In the following analysis we assume that each dimension of the random vector $\vx$ is independent, 
and that the variance in each dimension is 1. 
The general form of the forward noising step of a diffusion model can be described as scaling and adding noise, \ie 
\begin{align}\label{eq:noising}
\vx_t =  s(t) \vx_0 + s(t) \sigma(t) \vz,
\end{align}
where $\vz \sim \gauss{0}{\rmI}$ and $\vx_0$ is a sample drawn from data distribution. $s(t)$ and $\sigma(t)$ are user-defined coefficients. Here the coefficient on noise $\vz$ is parameterized as the product of $s(t)$ and $\sigma(t)$ so that $\sigma(t)$ represents the noise-to-signal ratio in $\vx_t$.

SMLD~\cite{song2019generative, song2020improved}, DDIM~\cite{ddim} and Karras~\cite{karras2022elucidating} 
sets scaling,~\ie $s(t) = 1$, and therefore 
adding noise by $\vx_0 + \sigma(t)\vz$ would cause $\vx_t$ to numerically get larger as $t$ increases. 
DDPM on the other hand introduced rapidly decreasing $s(t)$ to scale down the successive $\vx_t$ so that at any time $t$, $p_t(x)$ has variance fixed at 1. This goal of maintaining a standard deviation $1$ requires that 
\bea
\Var[\vx_t] = \Var \qty[ s(t) \vx_0 ] + \Var \qty[ s(t) \sigma(t) \vz ] \\
 s(t)^2 \underbrace{\Var[\vx_0]}_{=\rmI} + s(t)^2 \sigma(t)^2 \underbrace{ \Var[\vz] }_{=\rmI} = \rmI \\
s(t)^2  + s(t)^2 \sigma(t)^2  = 1 \\
\sigma(t) = \sqrt{\frac{1 - s(t)^2}{s(t)^2}}
\eea

DDPM specifies the $s(t)$ by a set of $\beta_t$, \ie, $s(t) = \sqrt{\bar{\alpha}_t} = \sqrt{ \prod_{i \le t} \alpha_i} = \sqrt{ \prod_{i \le t} (1 - \beta_i)}$ , and therefore $\sigma(t) = \sqrt{\frac{1 - \bar{\alpha}_t}{\bar{\alpha}_t}}$.

The noising step \eqref{eq:noising} describes the marginal distribution at $p_t(\vx)$. The infinitesimal time 
evolution of this process can be written as the following stochastic differential equation~\cite{song2021score}: 
\begin{align} \label{eq:sde}
\dd{\vx} &= f(t) \vx \dd{t} + g(t) \dd{\vomega_t} \quad \text{where} \quad f(t) = \frac{\dot{s}(t)}{s(t)} \quad g(t) = s(t) \sqrt{2 \dot{\sigma}(t) \sigma(t) }. 
\end{align}
Fokker-Planck equation~\cite{fokker-planck} states that a stochastic differential equation of the form \eqref{eq:sde} is identified with a partial differential equation describing the marginal probability density distribution $p_t(\vx)$
\begin{align}
\dd{ \vx } &= f(x, t) \dd{t} + g(t) \dd{ \vomega_t} ~\longleftrightarrow~ \pdv{p_t(\vx)}{t} = - \div \qty[ f(x, t) \, p_t(\vx) - \frac{g(t)^2}{2} \grad_x p_t(\vx) ]. 
\end{align}
Applying this identity tells us that a stochastic differential equation like \eqref{eq:sde} implies 
a deterministic, ordinary differential equation. Here we illustrate the proof schematically: 
\begin{eqnarray}
\begin{tikzcd}
\underbrace{ 
    \dd{ \vx } = \textcolor{blue}{ f(t) \, \vx} \dd{t} + \textcolor{red}{ g(t) } \dd{ \vomega_t}
}_{\text{stochastic}}
\arrow[r, Rightarrow, "\text{FP}"]
\arrow[d, dashed, "\text{implies}"] & 
\pdv{p_t(\vx)}{t} = - \div \qty[ \textcolor{blue}{ f(t) \, \vx \,} p_t(\vx) - \textcolor{red}{ \frac{g(t)^2}{2} } \grad_x p_t(\vx) ] 
\arrow[d, Rightarrow, "\text{equal (by log derivative trick; expanded below)}"] \\
\underbrace{ 
    \dd{\vx} = \scalemath{0.8}{ \textcolor{blue}{ \qty( f(t) \, \vx - \frac{g(t)^2}{2} \grad_{\vx} \log p_t(\vx) ) } } \dd{t} + \textcolor{red}{0} \dd{\vomega_t} 
}_{\text{deterministic}} & 
\pdv{p_t(\vx)}{t}  = - \div \qty[  
\scalemath{0.8}{\textcolor{blue}{ \qty( f(t) \, \vx - \frac{g(t)^2}{2} \grad_{\vx} \log p_t(\vx) ) } } 
p_t(\vx) - \textcolor{red}{0} ] 
\arrow[l, Rightarrow, "\text{FP}"]
\end{tikzcd}
\end{eqnarray}
The application of the log derivative trick is expanded below: 
\begin{align}
\pdv{p_t(\vx)}{t} 
&= - \div \qty[ f(t) \, \vx \, p_t(\vx) - \frac{g(t)^2}{2} \grad_x p_t(\vx) ] \\
&= - \div \qty[ \frac{ f(t) \, \vx \, p_t(\vx) - \frac{g(t)^2}{2} \grad_x p_t(\vx) }{p_t(\vx)} p_t(\vx) ] \\
&= - \div \qty[ \qty( f(t) \, \vx - \frac{g(t)^2}{2} \underbrace{ \frac{\grad_x p_t(\vx)}{p_t(\vx)} }_{\text{log derivative}} ) p_t(\vx) ] \\
&= - \div \qty[ \qty( f(t) \, \vx - \frac{g(t)^2}{2} \grad_{\vx} \log p_t(\vx) ) p_t(\vx) ].
\end{align}

Substituting the expression for $f(t)$ and $g(t)$ from \eqref{eq:sde}, 
we obtain an ODE from which we can sample the data by applying the score function with a step 
schedule that theoretically guarantees to take us back to initial, clean data distribution
\begin{align}
\dd{\vx} &= \frac{\dot{s}(t)}{s(t)} \vx - \frac{1}{2} \left( s(t) \sqrt{2 \dot{\sigma}(t) \sigma(t)} \right)^2 \grad_{\vx}\log p_t(\vx)  \dd{t} \\
\dd{\vx} &= \frac{\dot{s}(t)}{s(t)} \vx - s(t) \frac{\dot{\sigma}(t)}{\sigma(t)} \bigg( D(\vx / s(t); \sigma(t)) - \vx / s(t) \bigg) \dd{t}.
\end{align}
When $s(t) = 1$, $\sigma(t) = t$, the above simplifies to 
\begin{align}
\dd{\vx} &= - \sigma_t \cdot \frac{D(\vx; \sigma_t) - \vx}{\sigma_t^2} \dd{t} \\
\dd{\vx} &= - \sigma_t \cdot \text{score}(\vx, \sigma_t) \dd{t} \label{eq:final_ode}
\end{align}
Note that this schedule with $s(t) = 1$, $\sigma(t) = t$ allows for taking large step sizes during inference since it introduces no extra curvature in the trajectory beyond what's induced by the score function itself. 
The discretized sampling algorithm of equation~\eqref{eq:final_ode} is described in~\figref{fig:score_algo}.

%% file: 07_supp_results.tex
\section{Additional Experiments}\label{sec:more_results}
\input{figures/depth_ablations}
{\bf\noindent Ablation on center depth loss.}
In~\cref{fig:depth_ablations}, we illustrate the effect of the center depth loss proposed in~\equref{eq:center_depth_loss}. Without the center depth loss, we observe that some objects, \eg, French Fries, are placed far from the center of the scene box and tend to drift around when the camera
viewpoints are changed. This effect is more pronounced in the provided video result.
In contrast, a moderate center depth loss forces the object to be placed at the scene box center. Additionally, we observe that
the objects tend to be enlarged to occupy more of the visible screen space without wasting model capacity.

\input{figures/fig_supp_results.tex}
\medskip
{\bf\noindent Additional qualitative results.}
We provide additional qualitative results from SJC in~\cref{fig:supp_results}. Note that we increase the resolution 
of the depth maps beyond the $64 \times 64$ resolution of the image latents by rendering subpixel rays. In general, we observe
that the volumetric renderer is powerful enough to hallucinate shadows (horse), water surfaces (Sydney opera house, duck), grasslands (zebra) and even a traffic lane (school bus), using the volume densities.

\medskip
{\bf\noindent Video results.}
We have attached numerous video results in the supplemental materials. Please see the attached HTML and videos. %
We have named each file after the text prompt used to generate the 3D asset. In addition, we included the videos
for the ablation experiments in~\figref{fig:emptiness_ablations} and~\figref{fig:depth_ablations}. %

%% file: figures/depth_ablations.tex
\begin{figure}[t]
\centering
\setlength{\tabcolsep}{0pt}
\begin{tabular}{ccc@{\hspace{0.3em}}ccc}

\multicolumn{6}{c}{\small A high quality photo of french fries from McDonald's} \\
 \includegraphics[width = .16\textwidth]{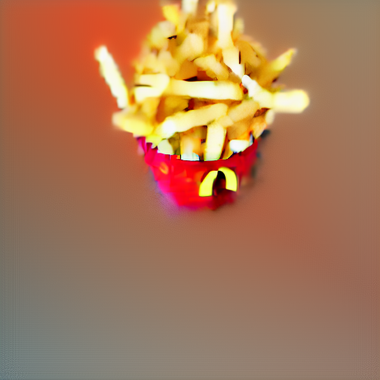}&
 \includegraphics[width = .16\textwidth]{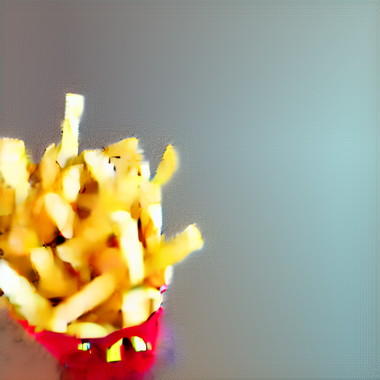}&
 \includegraphics[width = .16\textwidth]{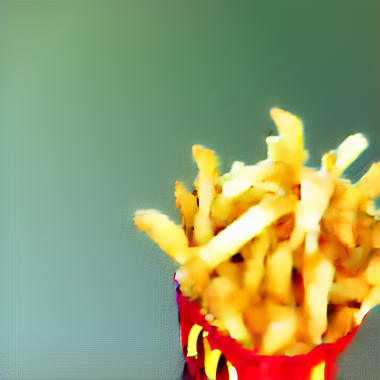}&
 \includegraphics[width = .16\textwidth]{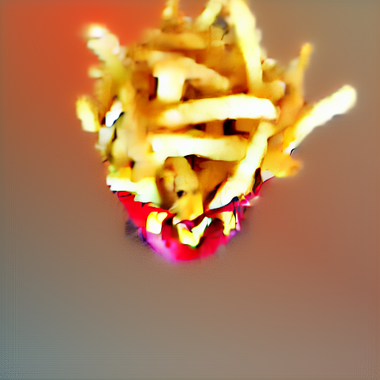}&
 \includegraphics[width = .16\textwidth]{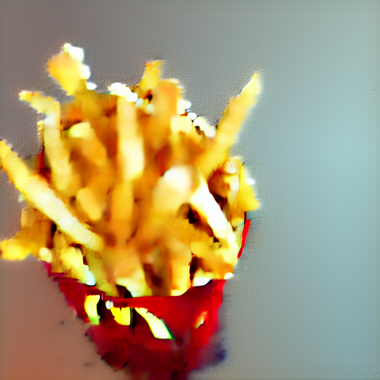}&
 \includegraphics[width = .16\textwidth]{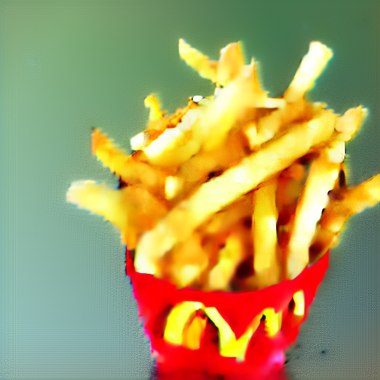}\\

\multicolumn{6}{c}{\small a DSLR photo of a rose} \\
 \includegraphics[width = .16\textwidth]{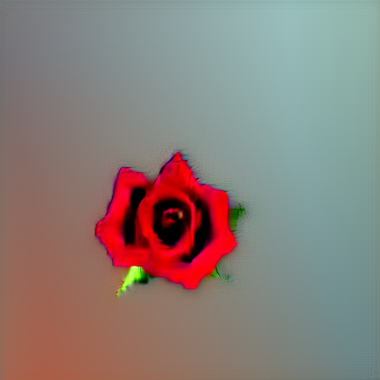}&
 \includegraphics[width = .16\textwidth]{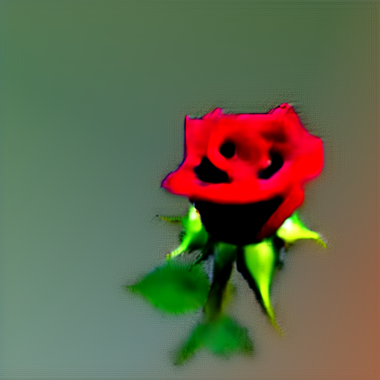}&
 \includegraphics[width = .16\textwidth]{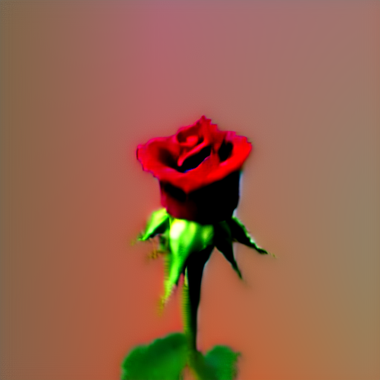}&
 \includegraphics[width = .16\textwidth]{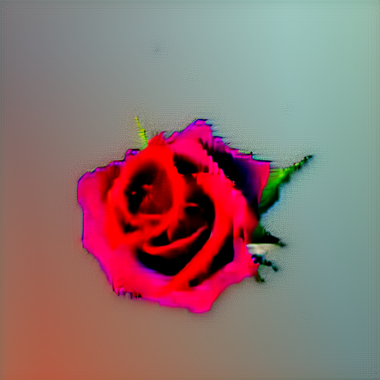}&
 \includegraphics[width = .16\textwidth]{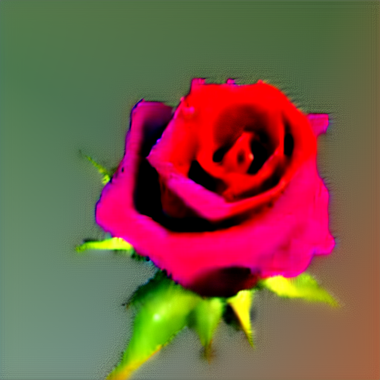}&
 \includegraphics[width = .16\textwidth]{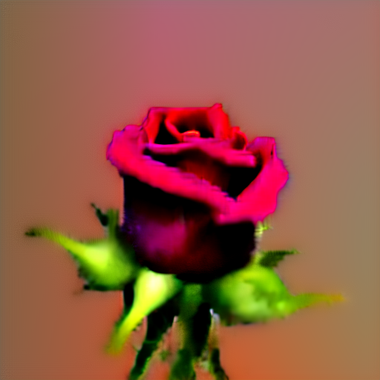}\\

\multicolumn{3}{c}{\small No center depth loss} & \multicolumn{3}{c}{\small With center depth loss (weight = 100)}  
\end{tabular}
\caption{Ablation experiments on the proposed center depth loss. Each pair of corresponding columns of the same prompt are visualized from the same camera angle. 
}
\vspace{-0.4cm}
\label{fig:depth_ablations}
\end{figure}

%% file: figures/fig_supp_results.tex
\begin{figure*}[!htb]
    \centering
    \setlength{\tabcolsep}{2pt}
    \renewcommand{\arraystretch}{0.95}
    \begin{tabular}{cccc}
    
    \multicolumn{4}{c}{ \small \begin{tabular}{c}
    Trump figure
    \end{tabular}} \\
    \includegraphics[width=0.24\linewidth]{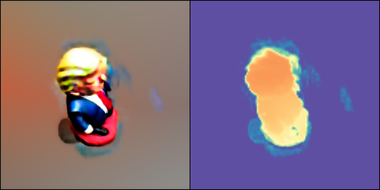} & 
    \includegraphics[width=0.24\linewidth]{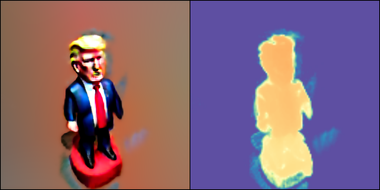} & 
    \includegraphics[width=0.24\linewidth]{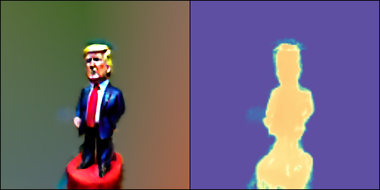} & 
    \includegraphics[width=0.24\linewidth]{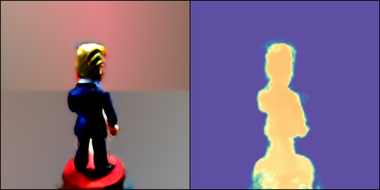}\\

    \multicolumn{4}{c}{ \small \begin{tabular}{c}
    Obama figure
    \end{tabular}} \\
    \includegraphics[width=0.24\linewidth]{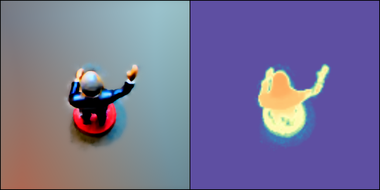} & 
    \includegraphics[width=0.24\linewidth]{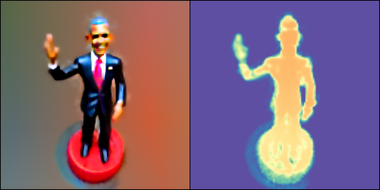} & 
    \includegraphics[width=0.24\linewidth]{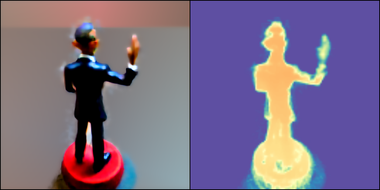} & 
    \includegraphics[width=0.24\linewidth]{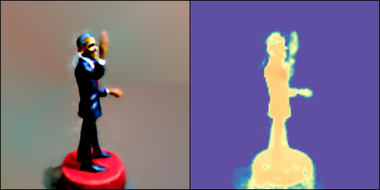}\\
    
    \multicolumn{4}{c}{ \small \begin{tabular}{c}
    Biden figure
    \end{tabular}} \\
    \includegraphics[width=0.24\linewidth]{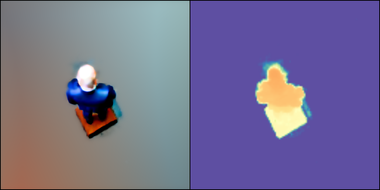} & 
    \includegraphics[width=0.24\linewidth]{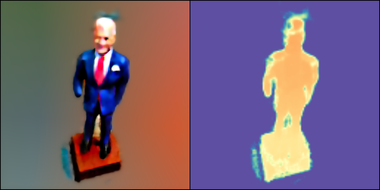} & 
    \includegraphics[width=0.24\linewidth]{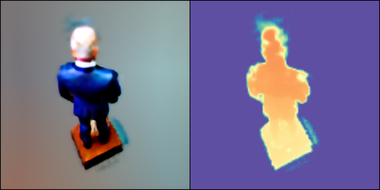} & 
    \includegraphics[width=0.24\linewidth]{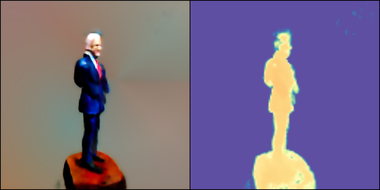}\\

    \multicolumn{4}{c}{ \small \begin{tabular}{c}
    Zelda Link
    \end{tabular}} \\
    \includegraphics[width=0.24\linewidth]{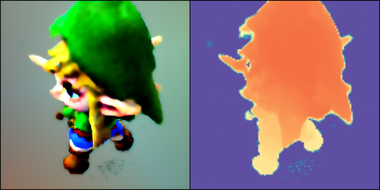} & 
    \includegraphics[width=0.24\linewidth]{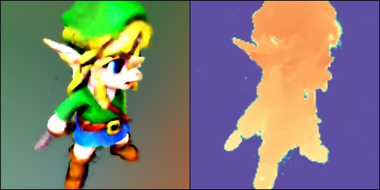} & 
    \includegraphics[width=0.24\linewidth]{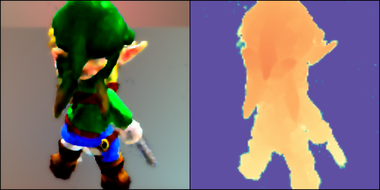} & 
    \includegraphics[width=0.24\linewidth]{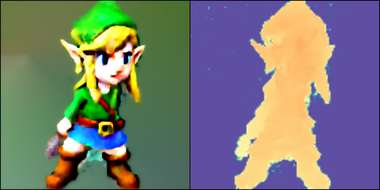}\\

    \multicolumn{4}{c}{ \small \begin{tabular}{c}
    A product photo of a Canon home printer
    \end{tabular}} \\
    \includegraphics[width=0.24\linewidth]{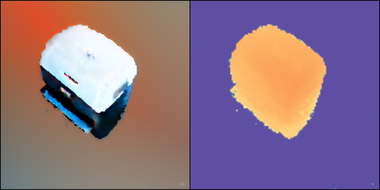} & 
    \includegraphics[width=0.24\linewidth]{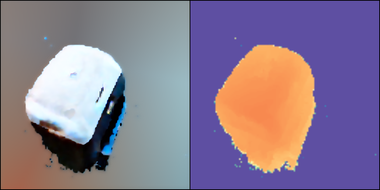} & 
    \includegraphics[width=0.24\linewidth]{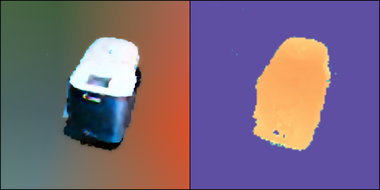} & 
    \includegraphics[width=0.24\linewidth]{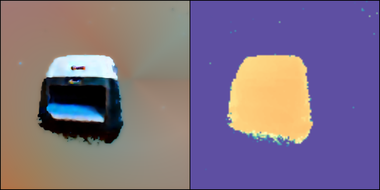}\\

    \multicolumn{4}{c}{ \small \begin{tabular}{c}
    A pig
    \end{tabular}} \\
    \includegraphics[width=0.24\linewidth]{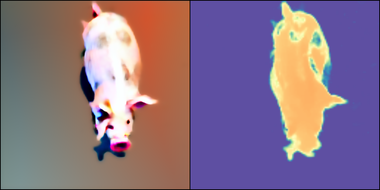} & 
    \includegraphics[width=0.24\linewidth]{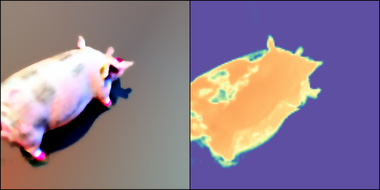} & 
    \includegraphics[width=0.24\linewidth]{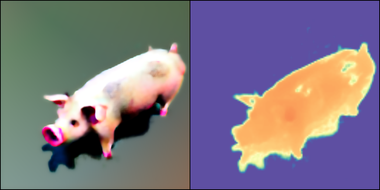} & 
    \includegraphics[width=0.24\linewidth]{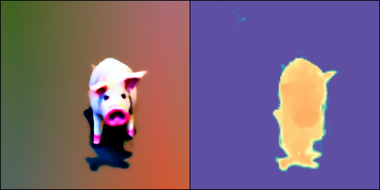}\\

    \multicolumn{4}{c}{ \small \begin{tabular}{c}
    A photo of a zebra walking
    \end{tabular}} \\
    \includegraphics[width=0.24\linewidth]{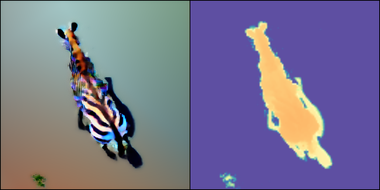} & 
    \includegraphics[width=0.24\linewidth]{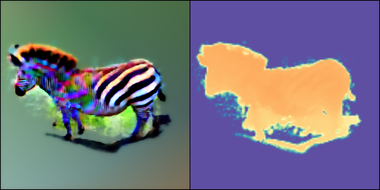} & 
    \includegraphics[width=0.24\linewidth]{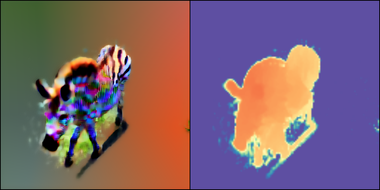} & 
    \includegraphics[width=0.24\linewidth]{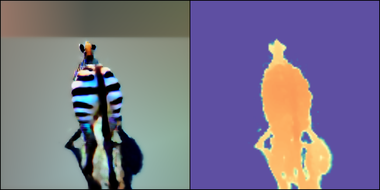}\\

    \multicolumn{4}{c}{ \small \begin{tabular}{c}
    A wide angle zoomed out photo of Saturn V rocket from distance
    \end{tabular}} \\
    \includegraphics[width=0.24\linewidth]{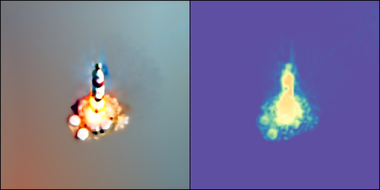} & 
    \includegraphics[width=0.24\linewidth]{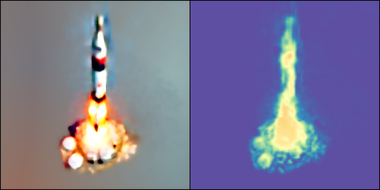} & 
    \includegraphics[width=0.24\linewidth]{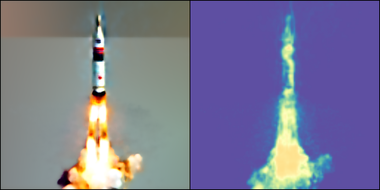} & 
    \includegraphics[width=0.24\linewidth]{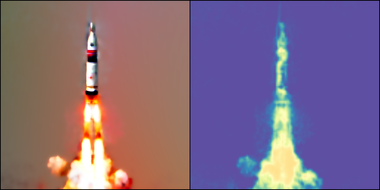}\\

    \multicolumn{4}{c}{ \small \begin{tabular}{c}
    A high quality photo of a yellow school bus
    \end{tabular}} \\
    \includegraphics[width=0.24\linewidth]{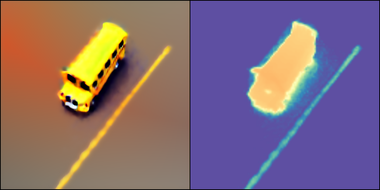} & 
    \includegraphics[width=0.24\linewidth]{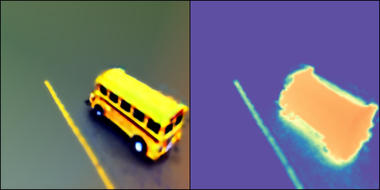} & 
    \includegraphics[width=0.24\linewidth]{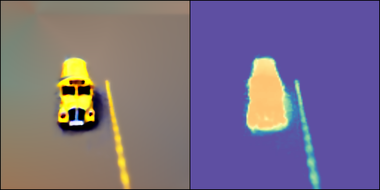} & 
    \includegraphics[width=0.24\linewidth]{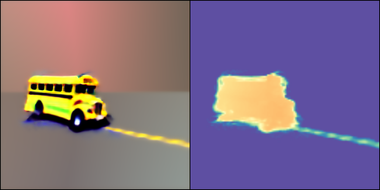}\\
    
    \end{tabular}
    \vspace{-0.4cm}
    \caption{Additional results of text-prompted generation of 3D models with SJC.
    }
    \vspace{-0.25cm}
    \label{fig:supp_results}
\end{figure*}

%% file: 07_supp_impl.tex
\section{Implementation Details}
\label{sec:impl_details}
{\noindent \bf 3D scene setup.} Our voxel grids are of size $100^3$, and placed
at world origin with a normalized side length $[-1, 1]^3$. We sample cameras uniformly on a hemisphere that covers the voxel cube with a radius of $1.5$, with look-at directions pointing at the origin. The camera field of view is randomly sampled from 40 degrees to 70 degrees during optimization, and fixed to 60 degrees at test time. We found the jittering on FoV to help with 3D optimization in some cases, and this data augmentation technique is reported in DreamFusion~\cite{dreamfusion}. 
Our scene background consists of an optimizable image of size $4 \times 4$ environment-mapped to the spherical surface by azimuth and elevation angles of the incoming ray. The small image size with constrained capacity helps to avoid confounding visual artifacts accumulating in the background during optimization. 

\medskip
{\noindent \bf Optimization.}
We use Adamax~\cite{adam} optimizer and perform gradient descent at a learning rate of $0.05$ for $10,000$ steps, with some prompts running at a longer schedule for better quality. Note that when performing gradient descent with NEScore, we implicitly rely on the optimizer's momentum state to perform the averaging. We have tried explicitly 
averaging the scores at multiple noise perturbations, but observed no benefits or degradation. 
The language-guidance scale is set to $100$. Our system consumes $9$GB of GPU memory during optimization, and takes approximately 25 minutes on an A6000 GPU including the time spent on miscellaneous tasks like visualization.

\medskip
{\noindent \bf View-dependent prompting.} 
An influence of DreamFusion~\cite{dreamfusion} on our work is the use of view-dependent prompting. 
Language prompts are prepended with one of the following: ``overhead view of'', ``front view of'', ``backside view of'', ``side view of'' depending on the camera location. More specifically, when camera elevation is above 30 degrees we use the ``overhead view'' prompt. Otherwise, the prompts are assigned based on the azimuth quadrant the camera falls into. This technique helps to alleviate the degeneracy of multiple frontal faces being painted around an object during optimization. We hope as part of our future work to develop a more general solution to induce the optimization towards more plausible geometry without %
using language as guidance. %

%% file: main.bbl
\begin{thebibliography}{10}\itemsep=-1pt

\bibitem[Baranchuk et~al.(2021)]{baranchuk2021label}
Dmitry Baranchuk, Ivan Rubachev, Andrey Voynov, Valentin Khrulkov, and Artem
  Babenko.
\newblock Label-efficient semantic segmentation with diffusion models.
\newblock {\em arXiv preprint arXiv:2112.03126}, 2021.

\bibitem[Cai et~al.(2020)]{shapegf}
Ruojin Cai, Guandao Yang, Hadar Averbuch-Elor, Zekun Hao, Serge Belongie, Noah
  Snavely, and Bharath Hariharan.
\newblock Learning gradient fields for shape generation.
\newblock In {\em Eur. Conf. Comput. Vis.}, 2020.

\bibitem[Chan et~al.(2022)]{eg3d}
Eric~R Chan, Connor~Z Lin, Matthew~A Chan, Koki Nagano, Boxiao Pan, Shalini
  De~Mello, Orazio Gallo, Leonidas~J Guibas, Jonathan Tremblay, Sameh Khamis,
  et~al.
\newblock Efficient geometry-aware {{3D}} generative adversarial networks.
\newblock In {\em IEEE Conf. Comput. Vis. Pattern Recog.}, 2022.

\bibitem[Chan et~al.(2021)]{pigan}
Eric~R Chan, Marco Monteiro, Petr Kellnhofer, Jiajun Wu, and Gordon Wetzstein.
\newblock pi-{GAN}: Periodic implicit generative adversarial networks for
  3d-aware image synthesis.
\newblock In {\em IEEE Conf. Comput. Vis. Pattern Recog.}, 2021.

\bibitem[Chang et~al.(2015)]{chang2015shapenet}
Angel~X Chang, Thomas Funkhouser, Leonidas Guibas, Pat Hanrahan, Qixing Huang,
  Zimo Li, Silvio Savarese, Manolis Savva, Shuran Song, Hao Su, et~al.
\newblock {ShapeNet}: An information-rich {3D} model repository.
\newblock {\em arXiv preprint arXiv:1512.03012}, 2015.

\bibitem[Chen et~al.(2022)]{tensorf}
Anpei Chen, Zexiang Xu, Andreas Geiger, Jingyi Yu, and Hao Su.
\newblock Tensorf: Tensorial radiance fields.
\newblock {\em arXiv preprint arXiv:2203.09517}, 2022.

\bibitem[Cheng(1995)]{meanshift}
Yizong Cheng.
\newblock Mean shift, mode seeking, and clustering.
\newblock {\em IEEE Trans. Pattern Anal. Mach. Intell.}, 1995.

\bibitem[Comaniciu and Meer(1999)]{mean-shift}
Dorin Comaniciu and Peter Meer.
\newblock Mean shift analysis and applications.
\newblock In {\em Int. Conf. Comput. Vis.}, 1999.

\bibitem[Deltombe(2022)]{deltombe_2022}
Amélie Deltombe.
\newblock How much does it cost to create {3D} models?, Apr 2022.

\bibitem[Dhariwal and Nichol(2021)]{dbeatgan}
Prafulla Dhariwal and Alexander Nichol.
\newblock Diffusion models beat {GANs} on image synthesis.
\newblock In {\em Adv. Neural Inform. Process. Syst.}, 2021.

\bibitem[Graikos et~al.(2022)]{ppprior}
Alexandros Graikos, Nikolay Malkin, Nebojsa Jojic, and Dimitris Samaras.
\newblock Diffusion models as plug-and-play priors.
\newblock {\em arXiv preprint arXiv:2206.09012}, 2022.

\bibitem[Ho et~al.(2020)]{ho2020denoising}
Jonathan Ho, Ajay Jain, and Pieter Abbeel.
\newblock Denoising diffusion probabilistic models.
\newblock In {\em Adv. Neural Inform. Process. Syst.}, 2020.

\bibitem[Ho and Salimans(2022)]{ho_guidance}
Jonathan Ho and Tim Salimans.
\newblock Classifier-free diffusion guidance.
\newblock {\em arXiv preprint arXiv:2207.12598}, 2022.

\bibitem[Hong et~al.(2022)]{avatarclip}
Fangzhou Hong, Mingyuan Zhang, Liang Pan, Zhongang Cai, Lei Yang, and Ziwei
  Liu.
\newblock Avatarclip: Zero-shot text-driven generation and animation of {3D}
  avatars.
\newblock {\em arXiv preprint arXiv:2205.08535}, 2022.

\bibitem[Hyv{\"a}rinen and Dayan(2005)]{hyvarinen2005estimation}
Aapo Hyv{\"a}rinen and Peter Dayan.
\newblock Estimation of non-normalized statistical models by score matching.
\newblock {\em J. Mach. Learn. Res}, 2005.

\bibitem[Jain et~al.(2022)]{dreamfields}
Ajay Jain, Ben Mildenhall, Jonathan~T Barron, Pieter Abbeel, and Ben Poole.
\newblock Zero-shot text-guided object generation with dream fields.
\newblock In {\em IEEE Conf. Comput. Vis. Pattern Recog.}, 2022.

\bibitem[Jetchev(2021)]{jetchev2021clipmatrix}
Nikolay Jetchev.
\newblock Clipmatrix: Text-controlled creation of {3D} textured meshes.
\newblock {\em arXiv preprint arXiv:2109.12922}, 2021.

\bibitem[Karras et~al.(2022)]{karras2022elucidating}
Tero Karras, Miika Aittala, Timo Aila, and Samuli Laine.
\newblock Elucidating the design space of diffusion-based generative models.
\newblock {\em arXiv preprint arXiv:2206.00364}, 2022.

\bibitem[Karras et~al.(2019)]{stylegan}
Tero Karras, Samuli Laine, and Timo Aila.
\newblock A style-based generator architecture for generative adversarial
  networks.
\newblock In {\em IEEE Conf. Comput. Vis. Pattern Recog.}, 2019.

\bibitem[Khalid et~al.(2022)]{clipmesh}
Nasir Khalid, Tianhao Xie, Eugene Belilovsky, and Tiberiu Popa.
\newblock Clip-mesh: Generating textured meshes from text using pretrained
  image-text models.
\newblock {\em ACM Trans. Graph.}, 2022.

\bibitem[Kingma et~al.(2021)]{kingma2021variational}
Diederik Kingma, Tim Salimans, Ben Poole, and Jonathan Ho.
\newblock Variational diffusion models.
\newblock {\em Advances in neural information processing systems},
  34:21696--21707, 2021.

\bibitem[Kingma and Ba(2014)]{adam}
Diederik~P Kingma and Jimmy Ba.
\newblock Adam: A method for stochastic optimization.
\newblock {\em arXiv preprint arXiv:1412.6980}, 2014.

\bibitem[Koch et~al.(2019)]{koch2019abc}
Sebastian Koch, Albert Matveev, Zhongshi Jiang, Francis Williams, Alexey
  Artemov, Evgeny Burnaev, Marc Alexa, Denis Zorin, and Daniele Panozzo.
\newblock {{ABC}}: {{A}} big cad model dataset for geometric deep learning.
\newblock In {\em IEEE Conf. Comput. Vis. Pattern Recog.}, 2019.

\bibitem[Lassner and Zollhofer(2021)]{pulsar}
Christoph Lassner and Michael Zollhofer.
\newblock Pulsar: Efficient sphere-based neural rendering.
\newblock In {\em IEEE Conf. Comput. Vis. Pattern Recog.}, 2021.

\bibitem[Lee and Chang(2022)]{clipvrf}
Han-Hung Lee and Angel~X Chang.
\newblock Understanding pure clip guidance for voxel grid nerf models.
\newblock {\em arXiv preprint arXiv:2209.15172}, 2022.

\bibitem[Li et~al.(2018)]{li2018differentiable}
Tzu-Mao Li, Miika Aittala, Fr{\'e}do Durand, and Jaakko Lehtinen.
\newblock Differentiable {Monte Carlo} ray tracing through edge sampling.
\newblock {\em ACM Trans. Graph.}, 2018.

\bibitem[Liu et~al.(2020)]{nsvf}
Lingjie Liu, Jiatao Gu, Kyaw Zaw~Lin, Tat-Seng Chua, and Christian Theobalt.
\newblock Neural sparse voxel fields.
\newblock In {\em Adv. Neural Inform. Process. Syst.}, 2020.

\bibitem[Liu et~al.(2021)]{fusedream}
Xingchao Liu, Chengyue Gong, Lemeng Wu, Shujian Zhang, Hao Su, and Qiang Liu.
\newblock Fusedream: Training-free text-to-image generation with improved clip+
  {GAN} space optimization.
\newblock {\em arXiv preprint arXiv:2112.01573}, 2021.

\bibitem[Loper and Black(2014)]{opendr}
Matthew~M Loper and Michael~J Black.
\newblock {OpenDR}: An approximate differentiable renderer.
\newblock In {\em Eur. Conf. Comput. Vis.}, 2014.

\bibitem[Maoutsa et~al.(2020)]{fokker-planck}
Dimitra Maoutsa, Sebastian Reich, and Manfred Opper.
\newblock Interacting particle solutions of fokker--planck equations through
  gradient--log--density estimation.
\newblock {\em Entropy}, 2020.

\bibitem[Martin-Brualla et~al.(2021)]{nerfw}
Ricardo Martin-Brualla, Noha Radwan, Mehdi~SM Sajjadi, Jonathan~T Barron,
  Alexey Dosovitskiy, and Daniel Duckworth.
\newblock Nerf in the wild: Neural radiance fields for unconstrained photo
  collections.
\newblock In {\em IEEE Conf. Comput. Vis. Pattern Recog.}, 2021.

\bibitem[Max(1995)]{maxvol}
Nelson Max.
\newblock Optical models for direct volume rendering.
\newblock {\em IEEE Trans. Vis. Comput. Graph.}, 1995.

\bibitem[Michel et~al.(2022)]{michel2022text2mesh}
Oscar Michel, Roi {Bar-On}, Richard Liu, Sagie Benaim, and Rana Hanocka.
\newblock Text2mesh: {{Text-driven}} neural stylization for meshes.
\newblock In {\em IEEE Conf. Comput. Vis. Pattern Recog.}, 2022.

\bibitem[Mildenhall et~al.(2021)]{nerf}
Ben Mildenhall, Pratul~P Srinivasan, Matthew Tancik, Jonathan~T Barron, Ravi
  Ramamoorthi, and Ren Ng.
\newblock Nerf: Representing scenes as neural radiance fields for view
  synthesis.
\newblock {\em {Commun. ACM}}, 2021.

\bibitem[Nguyen-Phuoc et~al.(2019)]{nguyen2019hologan}
Thu Nguyen-Phuoc, Chuan Li, Lucas Theis, Christian Richardt, and Yong-Liang
  Yang.
\newblock {HoloGAN}: Unsupervised learning of {3D} representations from natural
  images.
\newblock In {\em Int. Conf. Comput. Vis.}, 2019.

\bibitem[Nichol et~al.(2021)]{glide}
Alex Nichol, Prafulla Dhariwal, Aditya Ramesh, Pranav Shyam, Pamela Mishkin,
  Bob McGrew, Ilya Sutskever, and Mark Chen.
\newblock Glide: Towards photorealistic image generation and editing with
  text-guided diffusion models.
\newblock {\em arXiv preprint arXiv:2112.10741}, 2021.

\bibitem[Niemeyer and Geiger(2021)]{campari}
Michael Niemeyer and Andreas Geiger.
\newblock Campari: Camera-aware decomposed generative neural radiance fields.
\newblock In {\em Int. Conf. 3DV}, 2021.

\bibitem[Niemeyer and Geiger(2021)]{giraffe}
Michael Niemeyer and Andreas Geiger.
\newblock Giraffe: Representing scenes as compositional generative neural
  feature fields.
\newblock In {\em IEEE Conf. Comput. Vis. Pattern Recog.}, 2021.

\bibitem[Nimier-David et~al.(2019)]{mitsuba2}
Merlin Nimier-David, Delio Vicini, Tizian Zeltner, and Wenzel Jakob.
\newblock Mitsuba 2: A retargetable forward and inverse renderer.
\newblock {\em ACM Trans. Graph.}, 2019.

\bibitem[Oechsle et~al.(2021)]{unisurf}
Michael Oechsle, Songyou Peng, and Andreas Geiger.
\newblock Unisurf: Unifying neural implicit surfaces and radiance fields for
  multi-view reconstruction.
\newblock In {\em Int. Conf. Comput. Vis.}, 2021.

\bibitem[Poole et~al.(2022)]{dreamfusion}
Ben Poole, Ajay Jain, Jonathan~T Barron, and Ben Mildenhall.
\newblock {{DreamFusion}}: {{Text-to-3D}} using {{2D}} diffusion.
\newblock {\em arXiv preprint arXiv:2209.14988}, 2022.

\bibitem[Radford et~al.(2021)]{clip}
Alec Radford, Jong~Wook Kim, Chris Hallacy, Aditya Ramesh, Gabriel Goh,
  Sandhini Agarwal, Girish Sastry, Amanda Askell, Pamela Mishkin, Jack Clark,
  et~al.
\newblock Learning transferable visual models from natural language
  supervision.
\newblock 2021.

\bibitem[Rajeswar et~al.(2018)]{pix2scene}
Sai Rajeswar, Fahim Mannan, Florian Golemo, David Vazquez, Derek
  Nowrouzezahrai, and Aaron Courville.
\newblock Pix2scene: Learning implicit 3d representations from images.
\newblock {\em openreview}, 2018.

\bibitem[Ramesh et~al.(2022)]{dalle2}
Aditya Ramesh, Prafulla Dhariwal, Alex Nichol, Casey Chu, and Mark Chen.
\newblock Hierarchical text-conditional image generation with clip latents.
\newblock {\em arXiv preprint arXiv:2204.06125}, 2022.

\bibitem[Rombach et~al.(2022)]{rombach2022high}
Robin Rombach, Andreas Blattmann, Dominik Lorenz, Patrick Esser, and Bj{\"o}rn
  Ommer.
\newblock High-resolution image synthesis with latent diffusion models.
\newblock In {\em IEEE Conf. Comput. Vis. Pattern Recog.}, 2022.

\bibitem[Saharia et~al.(2022)]{imagen}
Chitwan Saharia, William Chan, Saurabh Saxena, Lala Li, Jay Whang, Emily
  Denton, Seyed Kamyar~Seyed Ghasemipour, Burcu~Karagol Ayan, S~Sara Mahdavi,
  Rapha~Gontijo Lopes, et~al.
\newblock Photorealistic text-to-image diffusion models with deep language
  understanding.
\newblock {\em arXiv preprint arXiv:2205.11487}, 2022.

\bibitem[Schuhmann et~al.(2022)]{laion}
Christoph Schuhmann, Romain Beaumont, Richard Vencu, Cade Gordon, Ross
  Wightman, Mehdi Cherti, Theo Coombes, Aarush Katta, Clayton Mullis, Mitchell
  Wortsman, et~al.
\newblock {LAION-5B}: An open large-scale dataset for training next generation
  image-text models.
\newblock {\em arXiv preprint arXiv:2210.08402}, 2022.

\bibitem[Schwarz et~al.(2020)]{graf}
Katja Schwarz, Yiyi Liao, Michael Niemeyer, and Andreas Geiger.
\newblock {GRAF}: Generative radiance fields for {3D}-aware image synthesis.
\newblock In {\em Adv. Neural Inform. Process. Syst.}, 2020.

\bibitem[Sohl-Dickstein et~al.(2015)]{sohl2015deep}
Jascha Sohl-Dickstein, Eric Weiss, Niru Maheswaranathan, and Surya Ganguli.
\newblock Deep unsupervised learning using nonequilibrium thermodynamics.
\newblock In {\em International Conference on Machine Learning}, pages
  2256--2265. PMLR, 2015.

\bibitem[Song et~al.(2021)]{ddim}
Jiaming Song, Chenlin Meng, and Stefano Ermon.
\newblock Denoising diffusion implicit models.
\newblock In {\em Int. Conf. Learn. Represent.}, 2021.

\bibitem[Song and Ermon(2019)]{song2019generative}
Yang Song and Stefano Ermon.
\newblock Generative modeling by estimating gradients of the data distribution.
\newblock In {\em Adv. Neural Inform. Process. Syst.}, 2019.

\bibitem[Song and Ermon(2020)]{song2020improved}
Yang Song and Stefano Ermon.
\newblock Improved techniques for training score-based generative models.
\newblock In {\em Adv. Neural Inform. Process. Syst.}, 2020.

\bibitem[Song et~al.(2021)]{song2021score}
Yang Song, Jascha {Sohl-Dickstein}, Diederik~P Kingma, Abhishek Kumar, Stefano
  Ermon, and Ben Poole.
\newblock Score-based generative modeling through stochastic differential
  equations.
\newblock In {\em Int. Conf. Learn. Represent.}, 2021.

\bibitem[Sun et~al.(2022)]{dvgo}
Cheng Sun, Min Sun, and Hwann-Tzong Chen.
\newblock Direct voxel grid optimization: Super-fast convergence for radiance
  fields reconstruction.
\newblock In {\em IEEE Conf. Comput. Vis. Pattern Recog.}, 2022.

\bibitem[Sun et~al.(2018)]{sun2018pix3d}
Xingyuan Sun, Jiajun Wu, Xiuming Zhang, Zhoutong Zhang, Chengkai Zhang, Tianfan
  Xue, Joshua~B Tenenbaum, and William~T Freeman.
\newblock {Pix3D}: Dataset and methods for single-image {3D} shape modeling.
\newblock In {\em IEEE Conf. Comput. Vis. Pattern Recog.}, 2018.

\bibitem[Vincent(2011)]{vincent2011connection}
Pascal Vincent.
\newblock A connection between score matching and denoising autoencoders.
\newblock {\em Neural computation}, 2011.

\bibitem[Wang et~al.(2021)]{neus}
Peng Wang, Lingjie Liu, Yuan Liu, Christian Theobalt, Taku Komura, and Wenping
  Wang.
\newblock Neus: Learning neural implicit surfaces by volume rendering for
  multi-view reconstruction.
\newblock {\em arXiv preprint arXiv:2106.10689}, 2021.

\bibitem[Wu et~al.(2016)]{wu3dgan}
Jiajun Wu, Chengkai Zhang, Tianfan Xue, Bill Freeman, and Josh Tenenbaum.
\newblock Learning a probabilistic latent space of object shapes via {3D}
  generative-adversarial modeling.
\newblock In {\em Adv. Neural Inform. Process. Syst.}, 2016.

\bibitem[Wu et~al.(2015)]{modelnet40}
Zhirong Wu, Shuran Song, Aditya Khosla, Fisher Yu, Linguang Zhang, Xiaoou Tang,
  and Jianxiong Xiao.
\newblock {3D ShapeNets}: A deep representation for volumetric shapes.
\newblock In {\em IEEE Conf. Comput. Vis. Pattern Recog.}, 2015.

\bibitem[Yang et~al.(2019)]{pointflow}
Guandao Yang, Xun Huang, Zekun Hao, Ming-Yu Liu, Serge Belongie, and Bharath
  Hariharan.
\newblock {PointFlow}: {3D} point cloud generation with continuous normalizing
  flows.
\newblock In {\em Int. Conf. Comput. Vis.}, 2019.

\bibitem[Yariv et~al.(2021)]{volsdf}
Lior Yariv, Jiatao Gu, Yoni Kasten, and Yaron Lipman.
\newblock Volume rendering of neural implicit surfaces.
\newblock In {\em Adv. Neural Inform. Process. Syst.}, 2021.

\bibitem[Yu et~al.(2021)]{plenoxels}
Alex Yu, Sara Fridovich-Keil, Matthew Tancik, Qinhong Chen, Benjamin Recht, and
  Angjoo Kanazawa.
\newblock Plenoxels: Radiance fields without neural networks.
\newblock {\em arXiv preprint arXiv:2112.05131}, 2021.

\bibitem[Yu et~al.(2015)]{lsun}
Fisher Yu, Ari Seff, Yinda Zhang, Shuran Song, Thomas Funkhouser, and Jianxiong
  Xiao.
\newblock {LSUN}: Construction of a large-scale image dataset using deep
  learning with humans in the loop.
\newblock {\em arXiv preprint arXiv:1506.03365}, 2015.

\bibitem[Zhang et~al.(2020)]{zhang2020image}
Yuxuan Zhang, Wenzheng Chen, Huan Ling, Jun Gao, Yinan Zhang, Antonio Torralba,
  and Sanja Fidler.
\newblock Image {GANs} meet differentiable rendering for inverse graphics and
  interpretable {3D} neural rendering.
\newblock {\em arXiv preprint arXiv:2010.09125}, 2020.

\bibitem[Zhao et~al.(2022)]{zhao2022generative}
Xiaoming Zhao, Fangchang Ma, David G{\"u}era, Zhile Ren, Alexander~G Schwing,
  and Alex Colburn.
\newblock Generative multiplane images: {{Making}} a {{2D GAN 3D-Aware}}.
\newblock In {\em Eur. Conf. Comput. Vis.}, 2022.

\end{thebibliography}
